%% file: main.tex
\documentclass[11pt, a4paper, onecolumn, copyright, goog]{google}
\usepackage{subcaption}
\usepackage[authoryear, sort&compress, round]{natbib}
\usepackage{comment}
\usepackage{multirow}
\usepackage{hyperref}
\hypersetup{
    colorlinks=true,
    linkcolor=blue,
    urlcolor=blue,
}

\usepackage[utf8]{inputenc}
\usepackage{xcolor} 

\newcommand{\MedAstraName}{AI co-clinician}

\uselogo{} 

\title{Towards Conversational Medical AI \\with Eyes, Ears and a Voice}

\renewcommand{\today}{2026-04-28}

\author[*, 1]{Meet Shah}
\author[*, 3]{Jason Gusdorf}
\author[*, 2]{Anil Palepu}
\author[*, 1]{Chunjong Park}  
\author[4]{Jack W. O'Sullivan}
\author[4]{Vishnu Ravi}
\author[1]{\\Tim Strother}
\author[1]{Pavel Dubov}
\author[1]{Aliya Rysbek}
\author[1]{Toshiyuki Fukuzawa}
\author[1]{Yana Lunts}
\author[1]{Jan Freyberg}
\author[1]{\\Michael B. Chang}
\author[1]{Aniruddh Raghu}
\author[1]{David Stutz}
\author[1]{Devora Berlowitz}
\author[1]{Eliseo Papa}
\author[1]{Taylan Cemgil}
\author[1]{\\JD Velasquez}
\author[1]{Jack Chen}
\author[1]{Arthur Chen}
\author[1]{Doug Fritz}
\author[1]{Charlie Taylor}
\author[1]{Katya Tregubova}
\author[1]{Jing Rong Lim}
\author[1]{Richard Green}
\author[1]{Sara Mahdavi}
\author[2]{Mahvish Nagda}
\author[2]{Jihyeon Lee}
\author[2]{Craig Schiff}
\author[1]{Liviu Panait}
\author[1]{Sukhdeep Singh} 
\author[1]{Valentin Li\'evin}
\author[1]{David G.T. Barrett}
\author[1]{Hannah Gladman}
\author[1]{Anna Cupani}
\author[1]{Francesca Pietra}
\author[1]{Uchechi Okereke}
\author[1]{Katherine Tong}
\author[1]{Clemens Meyer}
\author[1]{Erwan Rolland}
\author[1]{Mili Sanwalka}
\author[2]{Michael D. Howell} 
\author[1]{Shixiang Shane Gu} 
\author[1]{Bibo Xu}
\author[4]{Euan A. Ashley}
\author[1]{S. M. Ali Eslami}
\author[1]{Gregory Wayne}
\author[1]{Pushmeet Kohli}
\author[$\dagger$,1]{Vivek Natarajan}
\author[$\dagger$,3]{\\Adam Rodman}
\author[$\dagger$,1]{Alan Karthikesalingam}
\author[*, $\dagger$, 1]{Ryutaro Tanno}

\affil[1]{Google DeepMind}
\affil[2]{Google Research}
\affil[3]{Beth Israel Deaconess Medical Center, Harvard Medical School}
\affil[4]{Stanford University}
\affil[*]{Equal contributions}
\affil[$\dagger$]{Equal leadership}

\correspondingauthor{
$\{$rtanno$,\,$alankarthi$, \,$meetshah$, \,$natviv$\}@google.com, \{$jgusdorf$,\,$arodman$\}$@bidmc.harvard.edu
}

\begin{abstract} 

The practice of medicine relies not only upon skillful dialogue but also on the nuanced exchange and interpretation of rich auditory and visual cues between doctors and patients. Building on the low-latency voice and video processing capabilities of Gemini, we introduce \MedAstraName{}, a first-of-its-kind conversational AI system that utilizes continuous streams of audio-visual data from live patient conversations to inform real-time diagnostic and management decisions. Its dual-agent architecture balances deep clinical reasoning with the low latency required for natural dialogue, driving significant performance gains.
To assess this multimodal system, we implemented a video-based interface emulating a real telemedicine consultation. We crafted 20 standardized outpatient scenarios requiring proactive real-time auditory and visual reasoning and designed ``TelePACES'' evaluation criteria alongside case-specific rubrics. In a randomized, interface-blinded, crossover simulation study ($n=120$ encounters) with 10 internal medicine residents serving as patient actors, we compared \MedAstraName{} with primary care physicians (PCPs), GPT-Realtime, and a baseline agent.
\MedAstraName{} approached PCPs in key TelePACES dimensions, including the quality of management plans and differential diagnosis, while significantly outperforming GPT-Realtime across all general criteria. While our agent demonstrated parity with PCPs in case-specific triage measures, physicians maintained a superior overall performance hierarchy in case-specific assessments. Although \MedAstraName{} marks a significant advance to real-time telemedical AI, gaps remain in physical examination and disease-specific reasoning. Our work shows that text-only approaches fail to capture the true challenges of medical consultation and suggests that high-stakes real-time diagnostic AI is most safely advanced in  collaborative/triadic models where AI can be a supportive co-clinician for doctors and patients. 
\end{abstract}

\begin{document}

\maketitle

\section{Introduction}
The practice of medicine requires not only the skillful elicitation of a patient's history, but also expert attendance to a rich tapestry of visual and auditory signs observed during a clinical encounter. While LLMs have rapidly progressed in text-based conversational aspects of diagnostic consultations \citep{McDuff2025, Tu2025}, they have been unable to integrate the essential real-time audiovisual nuances of in-person or telemedical communication between doctors and patients \citep{korom2025aibasedclinicaldecisionsupport, zeltzer2025comparison, palepu2025conversationalaidiseasemanagement, saab2025advancingconversationaldiagnosticai}. In telehealth, auditory and visual cues are a key driver of clinical insights for both patients and physicians. The majority of remote medical consultations are performed over video calls, rather than through telephone visits or chat-based interactions~\citep{lee2023telehealth}. For clinicians, evaluating physical appearance, behavior, posture and clinical signs is a fundamental component of history-taking and the physical examination \citep{Wright2022-me}. Patients may also prefer video-enabled telehealth visits over audio-only interactions for new concerns as they permit more detailed consultations including the ability to be examined \citep{Kruis2024-zo}. Since the COVID-19 pandemic accelerated the adoption of telemedicine, the ``guided physical examination'' has emerged as a core competency \citep{noronha2022telehealth} and frameworks such as the ``Telehealth Ten'' illustrate how many traditional exam maneuvers can be adapted to the virtual setting \citep{Benziger2020-nd, Tong2025-gn}. Furthermore, text-only interfaces present challenges to accessibility (for example in patients with poor sight, mobility, language barriers or literacy issues) and can exacerbate health inequities \citep{Choi2023-uc}. Digital literacy is already a major determinant of healthcare outcomes, and recent reviews indicate that patients with low literacy often struggle to use text-based tools effectively despite having positive views of their potential utility \citep{Arias_Lopez2023-qk, Moore2025-sj, wamala2025digital, lee2023telehealth}.

\begin{figure}[t]
    \centering
    \includegraphics[width=\textwidth]{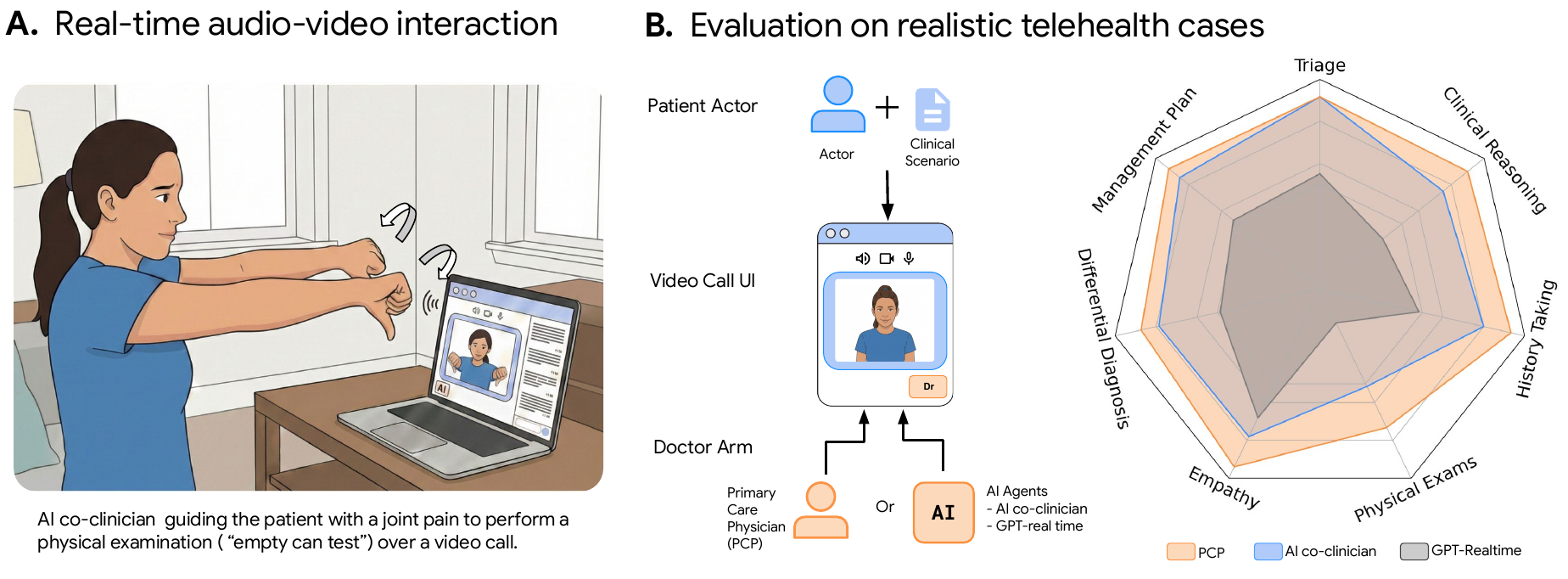}
    \caption{\footnotesize \textbf{Overview of key contributions.} \textbf{A.} Real-time telehealth interface where patient actors interact with AI agents or physicians via continuous audio-video streaming. \textbf{B.} Crossover comparative study: the performance of \MedAstraName{}, our audio-video medical agent was evaluated against 3 comparators including a pool of three primary care physicians (PCPs) and GPT-Realtime on 20 standardized clinical scenarios with interface-blinded patient actors. Performance assessment entailed the combination of case-specific rubrics across 7 domains and a universal rubric, enabling head-to-head comparison across history-taking, physical examination, clinical reasoning, communication, treatment, triage, and red-flag detection. }
    \label{fig:overview}
\end{figure}

The landscape of conversational AI has recently shifted with the introduction of native, low-latency multimodal streaming technologies, exemplified by Google's Project Astra, Gemini Live, and OpenAI's GPT-Realtime \citep{GDM_astra, OpenAIGPTRealtime}. These multimodal systems are capable of processing continuous audio and video streams, enabling full-duplex interactions with the approachable fluency of a video-call between human users. Despite the rapid maturation and adoption of these tools in consumer applications, their efficacy for clinical conversations is uncertain, especially in tasks such as live virtual consultations. The capabilities required for an effective consultation are broad and include three key areas: the passive recognition of clinical signs, such as the tone, prosody, or content of speech; the passive visual observation of gait, mood, faces, and dermatological signs; and the active administration of a guided physical examination, where the agent must direct patient maneuvers and synchronously interpret audiovisual findings. Prior research into the multimodal capabilities of conversational diagnostic AI systems has largely been limited to text-based chats that process static or pre-recorded multimodal clinical data~\citep{johri2025evaluation, saab2025advancingconversationaldiagnosticai}. In comparison, a live, continuous audio-visual agent for telemedical consultation with acceptably low-latency and capable real-time clinical reasoning, is a sizable technological leap. Such a system needs to simultaneously perceive dynamic visual and auditory cues specific to the medical domain, guide the conversation and participant in the physical world, reason clinically and converse empathetically in real time.

Alongside the technological complexity, real-time multimodal conversations also increase the difficulty of evaluation significantly. Text-based chatbots have been evaluated using global rubrics, derived from Objective Structured Clinical Examinations (OSCEs) for medical trainees \citep{Regehr1998-ai}. However, such evaluations lack considerable nuances because of the case-specific nature of an effective consultation. For example, a patient presenting with shortness of breath, weight gain, and orthopnea requires a different consultation than a patient presenting with three days of shortness of breath, cough, and fever, a nuance typically not captured in a global rubric. These challenges are even further compounded in evaluating audio-visual AI systems, which could manifest subtle errors of either omission or commission that can be clinically meaningful and require careful appraisal in the audio-visual domain. For example, these systems could misinterpret left-sided abdominal pain as occurring on the right; or miss clinically meaningful visual signs (for example, respiratory distress, anxiety or flushing) in ways that would not have been apparent in the evaluation of text-based AI systems. As AI systems undergo deployment in high-stakes clinical contexts, it is important to evaluate not only the global quality of consultations but also potentially safety-critical case-specific criteria, both to contextualize the relevance of a finding and to ensure it has not been incorrectly confabulated by an AI system. 

In order to advance conversational medical agents to meet the needs of patients and health systems, we introduce \MedAstraName{}, a first-of-its-kind conversational audio-video medical agent, and compare its performance in realistic simulated clinical encounters to other AI systems (GPT-Realtime and a baseline agent) and human primary care practitioners (PCPs). We introduce a novel framework for evaluating diagnostic telemedical consultations by audio-visual medical agents, combining both OSCE-inspired evaluations with case-specific rubrics designed specifically to evaluate evoked and unevoked physical exam signs. Our key contributions (Figure~\ref{fig:overview}) include:

\begin{itemize}
    \item \textbf{\MedAstraName{}}: We developed \MedAstraName{}, a real-time audio-visual conversational diagnostic system built on Gemini and Project Astra that decouples a fluent conversational interface from deep clinical reasoning to address the unique demands of live video-based consultations. By orchestrating two complementary agents—a ``Talker'' for low-latency, empathetic audio-visual interaction and a supervisory ``Clinical Planner'' for deep diagnostic reasoning, \MedAstraName{} achieves medical rigor without sacrificing conversational fluidity. 
    \item \textbf{Real-Time Audio-Video Evaluation Environment and Rubrics}: We implemented a realistic video-based UI to enable users to interact with \MedAstraName{} in a way that mirrors a typical remote consultation for virtual care. We developed evaluation rubrics, purpose-built for real-time audio-video clinical encounters and examined performance using 20 realistic outpatient scenarios that require nuanced and proactive real-time visual reasoning. Our rubrics objectively assess the extraction of both evoked and unevoked physical exam findings directly from patient actors.
    \item \textbf{Multi-Arm Multi-center Evaluation}: We conduct a multi-center randomized, interface-blinded crossover study comparing \MedAstraName{} against primary care physicians (PCPs), GPT-Realtime, and a baseline version of \MedAstraName{} without the planner (\MedAstraName{}-without-Planner). \MedAstraName{} outperformed GPT-Realtime and the baseline agent by a large margin across a range of consultation quality axes, corroborating the value of our architectural design. Orthogonally, this work presents the very first published rigorous evaluation of a real-time AV agent in any domain.
\end{itemize}


\section{Methods}

\begin{figure}[t]
    \centering
    \vspace{-3mm}
    \includegraphics[width=\textwidth]{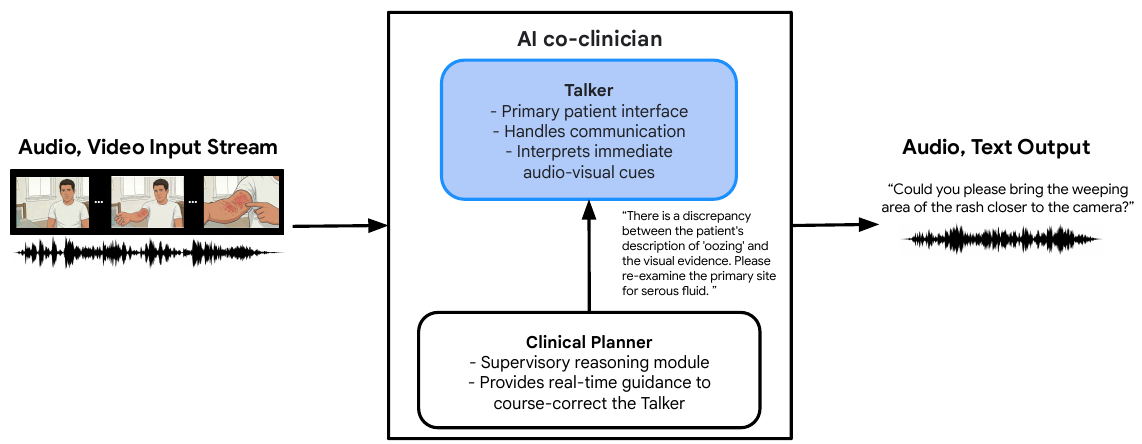}
     \caption{\footnotesize \textbf{Architecture overview of \MedAstraName{}.} The system comprises two complementary LLM-based agents: (1) the \textit{Talker}, which handles real-time patient interaction including audio generation, immediate audiovisual cue interpretation (e.g., respiratory effort, affect), and empathetic responsiveness; and (2) the \textit{Clinical Planner}, a supervisory reasoning module that tracks gathered symptoms, maintains clinical goals, synthesizes evidence against diagnostic protocols, and course-corrects the Talker to ensure systematic history-taking and exploration of differentials without sacrificing conversational fluidity.}
    \label{fig:agent_overview}
\end{figure}

\subsection{\MedAstraName{}}

\MedAstraName{} is an LLM-based system built on the Gemini family of models \citep{comanici2025gemini} designed for real-time diagnostic dialogue over a video-based interface. The system builds upon Gemini's native audio-visual understanding and Project Astra \citep{GDM_astra}, a framework for agentic orchestration and handling of bidirectional streaming. While the Astra framework provides the foundation for low-latency and fluid audio-visual interaction, clinical encounters impose a unique dual burden: 1) the need for immediate, empathetic responsiveness and 2) the requirement for structured, deep, methodical diagnostic reasoning. To optimize for this, \MedAstraName{} relies on the interplay of two complementary agents (Figure~\ref{fig:agent_overview}): the ``Talker'' and the ``Clinical Planner''. During the development phase, this ``separation of concerns'' enabled striking a better balance between fluency and deep clinical reasoning than having a single talker agent with adaptive reasoning due to the conflicting requirements of the two behaviors.

\noindent\textbf{Talker.} The Talker interfaces directly with the patient and is responsible for the immediate, low-level mechanics of the conversation—responding to the patient's questions, maintaining empathetic rapport, and interpreting audio-visual cues (e.g., respiratory effort, affect, visible distress) with minimal latency. Its primary objective is to sustain a natural, fluid dialogue while capturing clinically relevant observations in real time.

\noindent\textbf{Clinical Planner.} The Clinical Planner operates as a supervisory reasoning module, detached from the immediacy required for audio response generation. It maintains a running model of the clinical encounter—tracking gathered symptoms, active differentials, and outstanding clinical goals—and shares structured guidance to course-correct the Talker. This separation ensures that diagnostic reasoning is not diluted by conversational mechanics: the Planner synthesizes gathered evidence against clinical protocols and actively directs the Talker to address missing history, administer guided physical exams, or explore differentials, keeping the encounter medically rigorous without sacrificing conversational fluidity. Section~\ref{section:planner_examples} in Appendix provides qualitative examples of how the planner module assists the talker in different ways.

\noindent\textbf{Baseline Agent for Ablation.} In order to assess the effects of Clinical Planner, we also included a version of \MedAstraName{} that only utilizes the Talker Agent as a baseline in our expert evaluation, which we refer to as \MedAstraName{}-without-Planner.

\subsection{Patient Scenario Design}
Patient-facing telehealth agents must work in real time, processing live audio and video streams, rather than relying on clinical findings that have already been translated into text. In routine telehealth encounters, clinicians depend on visual inspection (e.g., rash pattern, work of breathing, visible distress), nonverbal behavior (e.g., affect, psychomotor slowing or agitation), and examination maneuvers performed by the patient under guidance (e.g., inhaler technique demonstration, range-of-motion testing, localization of pain, patient-assisted palpation). Text-based evaluations can overestimate performance by supplying these cues in narrative form and by avoiding the practical work of guiding a telehealth exam and communicating safely. Patients may also struggle to perform and translate guidance for physical exam maneuvers in traditional telehealth examinations, which would be considerably harder with a text-based interface \citep{Tong2024-fy}. Patients may also be unable to express their symptoms effectively in the form of text due to literacy, familiarity with digital tools, or many other socio-cultural, medical or other factors.

Grounding an evaluation of medical AI systems in telemedicine introduces specific challenges. Audiovisual signals are variable because of camera framing, lighting, background noise, and latency. Telehealth physical examination is constrained and requires the agent to choose feasible maneuvers, give clear stepwise instructions, confirm understanding, and safely interpret findings that may be incomplete. Standardized portrayals also vary across actors and sites, even with training. Safety assessment is a nuanced and central requirement, as agents must reliably and proactively decide when to screen for case-critical red flags that may not be volunteered, and escalate appropriately to ensure patient safety.

We addressed these challenges through careful scenario construction and rubric design. Scenario creation was guided by existing approaches to telehealth competencies and previously-published telehealth OSCEs \citep{aamc2021telehealth, Sartori2020-qr}.  Board-certified practicing physicians at two US academic medical centers (Beth Israel Deaconess Medical Center in Boston, MA, and Stanford University in Palo Alto, CA), with expertise in medical education and examination for remote care delivery, constructed standardized clinical scenarios to approximate common ambulatory and urgent-care telehealth presentations while preserving enough diagnostic ambiguity to require active hypothesis generation and triage.

Each scenario was developed from a case template that specified (1) a chief concern with a realistic onset and trajectory, (2) a limited set of initial, patient-available facts (symptoms, basic history, home vitals when appropriate), and (3) a predefined “ground truth” diagnosis and severity level, including a small number of high-stakes alternative diagnoses that could not be excluded without targeted questioning. To support consistent evaluation across model encounters, the case script specifically identified patient information that should only be volunteered if specifically or actively requested. These items were explicitly mapped to key decision points for red-flag detection, differential diagnosis, and next-step management. Patient scenarios also included telehealth-appropriate examination  findings, patient-performed maneuvers, and symptom provocation tests with scripted responses, as well as standardized escalation pathways (e.g., ED referral thresholds) to assess safety-critical judgment. All scenarios were iteratively refined through pilot enactments to eliminate unintended cues, align difficulty across cases, and ensure that essential findings were discoverable through clinically reasonable history and remote examination.

\subsection{Universal and Case-Specific Rubric Design}
Evaluating the clinical capabilities of real-time multimodal agents required the development of an evaluation harness. Text-based models have traditionally been assessed with global rubrics derived from OSCEs, the established framework for evaluating human clinicians. Through an iterative process, we adapted the Practical Assessment of Clinical Examination Skills (PACES) rubric to comport with the existing telehealth OSCE rubrics \citep{Sartori2020-qr}. This new telePACES rubric (Table~\ref{tab:universal_rubrics}) was used to score all encounters, providing consistent cross-scenario measurement tool for comparing overall encounter quality and safety behavior. 

We also developed case-specific rubrics using a consistent three step process: (1) domain specification (2) case-specific criteria within each domain; and (3) rater training and rubric calibration. 

We first defined a priori the core domains of clinical performance across scenarios:  (a) history taking, (b) telehealth physical examination, (c) clinical reasoning and synthesis, (d) communication and counseling, (e) red-flag identification and triage, and (f) treatment and next-step management. Domains were adapted from existing telehealth competencies and reflect the minimal set of competencies required for safe and effective outpatient or urgent-care telemedicine practice. 

Next, for each domain, we specified observable behaviors and outputs that could be assessed directly from the encounter transcript and video. These items were written to be (i) observable (verifiable from the recording only), (ii) clinically consequential (omission would plausibly change assessment, triage, or management), and (iii) compatible with multiple acceptable clinical styles. To balance standardization with clinical realism, we distinguished “non-negotiable” elements (critical safety steps such as identifying emergent features, recommending urgent evaluation when indicated, or avoiding contraindicated advice) from “flexible” elements (reasonable variations in phrasing, ordering, or depth of counseling). Each item was assigned an ordinal score from 0 to 2 with explicit anchors defining what constituted an absent, partially complete, or fully adequate performance; anchors were written in affirmative language and included examples of acceptable alternative expressions.

Finally, we piloted our scenarios with a trained patient actor and used this scoring experience to iteratively refine the rubrics. During this phase, we used structured error review to identify criteria that were ambiguous, redundant, or insufficiently sensitive to clinically important differences. These revisions focused on improving the specificity of criteria, reducing double counting across domains (for example, separating synthesis from counseling), and ensuring that safety-critical behaviors were captured even when delivered succinctly (for example, recognizing and asking about features such as fevers, chills and intravenous drug use were required ``red flags'' in a back pain case). After finalizing rubrics,  we developed a standardized scoring manual that included domain definitions, items, and examples. We then held a member checking session in which raters scored the same encounters and reconciled their scoring differences. The final rubric and scoring manual were fixed prior to conducting the study, and all encounters were evaluated using the same rubrics. 

\subsection{Study Design}

\paragraph{Overview and Oversight.} We conducted a randomized, interface-blinded, crossover simulation study comparing three multimodal large-language–model telehealth agents (\MedAstraName{}, GPT-realtime, and \MedAstraName{}-without-Planner) with a pool of three human primary care physician (PCP) comparators. All 20 scenarios developed by both academic centers were used among all groups.  A pool of 10 resident physicians from these centers portrayed the patients, enacting each scenario four times (once for each study arm, in a randomized order) \citep{barrows1993overview}, with 10 of the 20 scenarios replicated by an additional patient actor for all arms (Table~\ref{tab:resident_demographics}). Internal medicine resident physicians were selected because role-playing evoked (manifested only if proactively requested) and unevoked (manifested without being proactively requested) physical exam signs in the context of multiple disease states requires domain expertise. Encounters were performed between October 2025 and February 2026. The BIDMC institutional review board determined the study to be exempt; all role-playing participants signed a participation agreement permitting use of identifiable video data. No exclusion criteria were applied. Actors studied standardized training instructions and materials before participation and received technical support during the study period. Immediately after each encounter, each resident evaluated the performance of the agent or PCP using both a case-specific rubric (Table~\ref{tab:case_specific_rubrics}) and the universal rubric (Table~\ref{tab:universal_rubrics}).

\paragraph{Agents and Comparators.} The LLM agents were deployed as multimodal telehealth systems capable of real-time audio-video interaction within the same web-based interface. The comparator was one of three board-certified internal medicine attending physicians at both sites. PCPs were blinded to the scenario diagnosis and aware that performance was being compared with LLM-based agents. Agents had no access to the web, simulated electronic health records, or external clinical tools during encounters. All agents were evaluated in the same interface under identical audiovisual inputs. The GPT-Realtime \citep{OpenAIGPTRealtime} arm was powered by gpt-realtime-2025-08-28. Of note, GPT-Realtime does not natively support continuous real-time video ingestion; to enable a fair comparison, we approximated this capability by allowing the model to request image frames from the video stream via an asynchronous tool call, mirroring the frame-capture mechanism available to Astra. This represents the best available approximation for real-time visual input using GPT-Realtime's current public capabilities. The patient-facing telehealth interface remained identical across all arms—patients interacted with the same UI regardless of agent—and all other experimental conditions (audio streaming, encounter duration limits, rubric administration, and evaluator blinding) were held constant.

\paragraph{Randomization and Blinding.} For each patient actor, the order of encounters across the four arms (\MedAstraName{}, GPT-realtime, \MedAstraName{}-without-Planner, and PCP) was randomized. The telehealth interface was blinded such that patient actors were unaware of which LLM agent they were interacting with. Interactions with PCPs were not blinded. Study investigators reviewed recordings for audiovisual quality and to confirm adequate performance environment and adherence to standardized portrayal procedures.

\paragraph{Data Capture.} All encounters were conducted on laptops in quiet environments using a web-based telehealth interface accessed through Google Chrome. Encounters were screen-recorded and uploaded to a secure, access-restricted study folder for archival and quality assurance. Session identifiers and encounter metadata were recorded at the time of the encounter. Analyses were conducted on de-identified datasets.

\subsection{Statistical Analysis}

\paragraph{Scoring.} Performance was evaluated immediately after each encounter using both case-specific rubrics (Table~\ref{tab:case_specific_rubrics}) and TelePACES (Table~\ref{tab:universal_rubrics}). Case specific rubrics spanned seven domains: History-Taking, Physical Exam (Telehealth), Clinical Reasoning, Communication and Counseling, Treatment Steps, Triage, and Red Flags. Each domain contains multiple questions which were scored on a 0–2 scale (0 = poor, 1 = intermediate, 2 = excellent). Domain scores were calculated as the sum of questions within each domain, and the total case-specific score was the sum across domains (see Table~\ref{tab:case_specific_rubrics} for a concrete example). For criteria in the universal rubric, Likert scales were used to rate (from 1-5) the general quality of various aspects of history taking, communication, diagnosis, and management. Table~\ref{tab:universal_rubrics} describes the definitions of the respective criteria and the scores.

\paragraph{Qualitative Analysis.} After completion of data collection and scoring, we performed a qualitative study of all recordings using a team-based directed content analysis, with two parallel analytical targets; performance of the models themselves mapped to case-specific domains (with a large focus on the guided physical exams), and actor compliance, mapped directly to our instruction guide. Five study members (three clinicians, two non-clinicians) independently viewed eight videos that were selected for scoring heterogeneity. Group discussion developed initial themes from both analytical targets. The team members with clinical expertise then reviewed all videos and prepared a thematic analysis.

\paragraph{Statistical Analysis.} Analyses accounted for repeated measures arising from the crossover design. Normalized rubric scores and TelePACES scores were compared across conditions using ordinary least squares (OLS) regression. To isolate the effect of the study condition, models included the study arm as the primary independent variable, alongside fixed effects for the clinical scenario and patient actor to control for baseline differences in case difficulty and actor grading tendencies. Pairwise comparisons were performed between each LLM agent and the PCP comparator to assess the statistical significance of any observed differences.

\section{Results}
A total of 120 telehealth encounters were completed across the four study arms, corresponding to 30 encounters per arm. Participant demographics are shown in Table~\ref{tab:resident_demographics}. Four encounters were repeated due to technical difficulties.

\subsection{Global measures of consultation quality}

\begin{figure}[h]
    \centering
    \includegraphics[width=\textwidth]{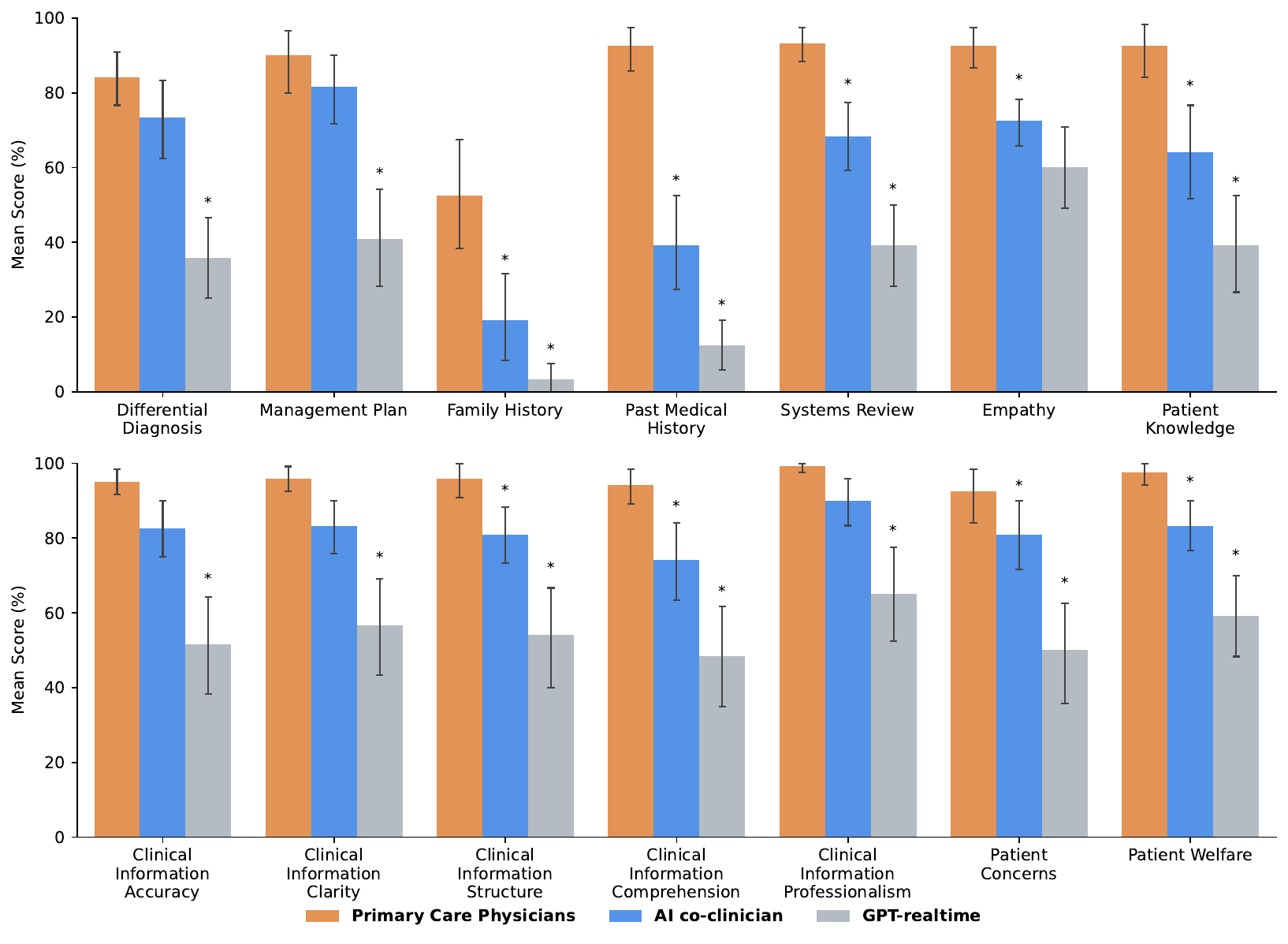}
    \caption{\footnotesize \textbf{Performance on universal evaluation rubrics (TelePACES).} Domain-agnostic criteria applied across all encounters, assessing communication quality (clarity, empathy, rapport), clinical reasoning process (differential generation, information synthesis), safety behaviors (red-flag screening, appropriate escalation), and overall consultation structure independent of case-specific content. For each criterion, Likert ratings from 1 to 5 were converted to a percent score and averaged across all scenarios. Error bars correspond to 95\% confidence intervals estimated based on $10^4$ bootstrap samples. Statistical significance ($* = p < 0.05$) was evaluated using ordinary least squares regression with fixed effects for clinical scenario and patient actor to compare 1) \MedAstraName{} to the Primary Care Physicians (indicated over the \MedAstraName{} bar) and 2) GPT-Realtime to \MedAstraName{} (indicated over the GPT-Realtime bar). Rubric details are provided in Table~\ref{tab:universal_rubrics}. }
    \label{fig:universal_rubric}
\end{figure}

\begin{figure}[t!]
    \centering
    \includegraphics[width=\textwidth]{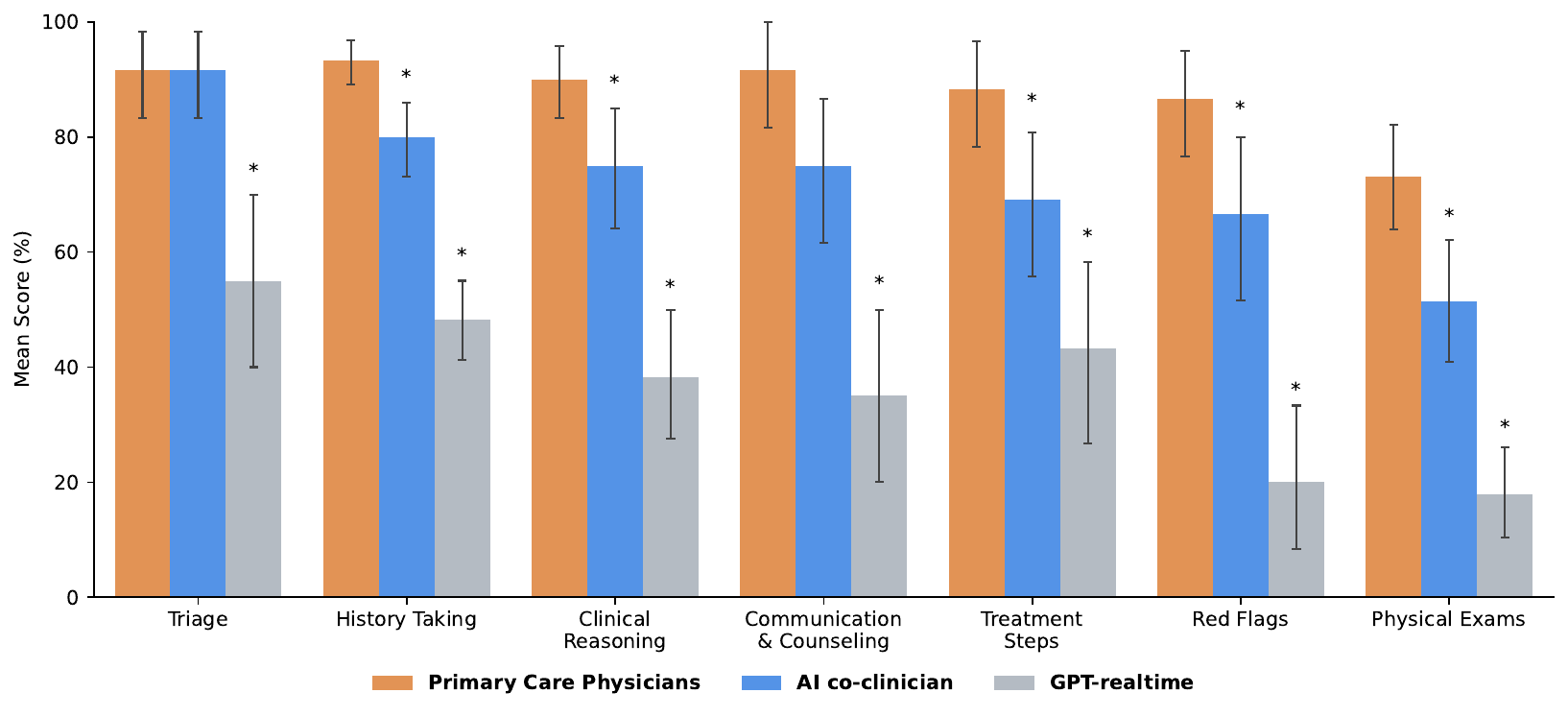}
    \caption{\footnotesize \textbf{Performance on case-specific evaluation rubrics.} Scenario-tailored criteria assessing seven domains: history-taking completeness, audiovisual cue extraction, patient-assisted physical examination (guided maneuvers and interpretation), clinical reasoning accuracy, communication and counseling, treatment recommendations, triage disposition, and red-flag detection. Each item uses anchored 0--2 scoring to distinguish omissions, partial completion, and fully appropriate performance. Error bars correspond to 95\% confidence intervals estimated based on $10^4$ bootstrap samples. Statistical significance ($* = p < 0.05$) was evaluated using ordinary least squares regression with fixed effects for clinical scenario and patient actor to compare 1) \MedAstraName{} to the Primary Care Physicians (indicated over the \MedAstraName{} bar) and 2) GPT-Realtime to \MedAstraName{} (indicated over the GPT-Realtime bar).}
    \label{fig:domain_results}
\end{figure}

Results from the global rubric are shown in Figure~\ref{fig:universal_rubric}. PCPs either outperformed or were equivalent to \MedAstraName{} in all domains, and consistently scored higher than GPT-realtime across the board. \MedAstraName{} approached physician-level performance in differential diagnosis (84.17\% (95\%CI, 76.67--91.67) for PCPs vs 73.33\% (95\%CI, 61.67--82.52) for \MedAstraName{}) as well as management plan quality (90.00\% (95\%CI, 80.83--96.67) for PCPs vs 81.67\% (95\%CI, 71.67--90.00) for \MedAstraName{}). On the other hand, the largest gaps were observed in measures of perceived empathy, the extent to which they confirmed patients' knowledge and understanding, and in eliciting a thorough past medical history or systems review. \MedAstraName{} significantly outperformed GPT-Realtime in 13 of 14 domains (except the empathy category).

\subsection{Case-specific measures of consultation quality}

Evaluation of the case-specific rubric domains showed a similar overall pattern as shown in Figure~\ref{fig:domain_results}. PCPs attained the highest mean case-specific rubric score of 85.34\% [95\%CI, 81.07--88.98] (see Table~\ref{tab:percentage_scores_mean_and_ci} in Appendix). \MedAstraName{} matched PCPs in the quality of triaging, with identical median scores of 91.67\% (95\%CI, 81.67–98.33), and was statistically non-inferior in communication and counseling (75.00\% [61.67--86.67] vs 91.67\% [81.67--100.00]). The largest differences between \MedAstraName{} and PCPs were in physical examination (51.47\% [40.47–62.53] vs. 73.19\% [64.30–81.47]), red-flag detection (66.67\% [51.67–80.00]) vs. 86.67\% [76.67–95.00]), and treatment steps (69.17\% [54.17–81.67] vs. 88.33\% [78.29–96.67]). \MedAstraName{} was the strongest AI system, significantly outperforming GPT-Realtime in every case-specific domain by large margins. In particular, largest discrepancies were observed in triage (91.67\% [IQR, 81.67–-98.33] vs. 55.00\% [40.00–68.33]), clinical reasoning (75.00\% [63.33–84.17] vs. 38.33\% [25.83–50.00]), red-flag detection (66.67\% [51.67–80.00] vs. 20.00\% [8.33–31.71]), and physical examination (51.47\% [40.47–62.53] vs. 17.83\% [10.42–25.70]). 

\begin{figure}[t]
    \centering
    \includegraphics[width=\textwidth]{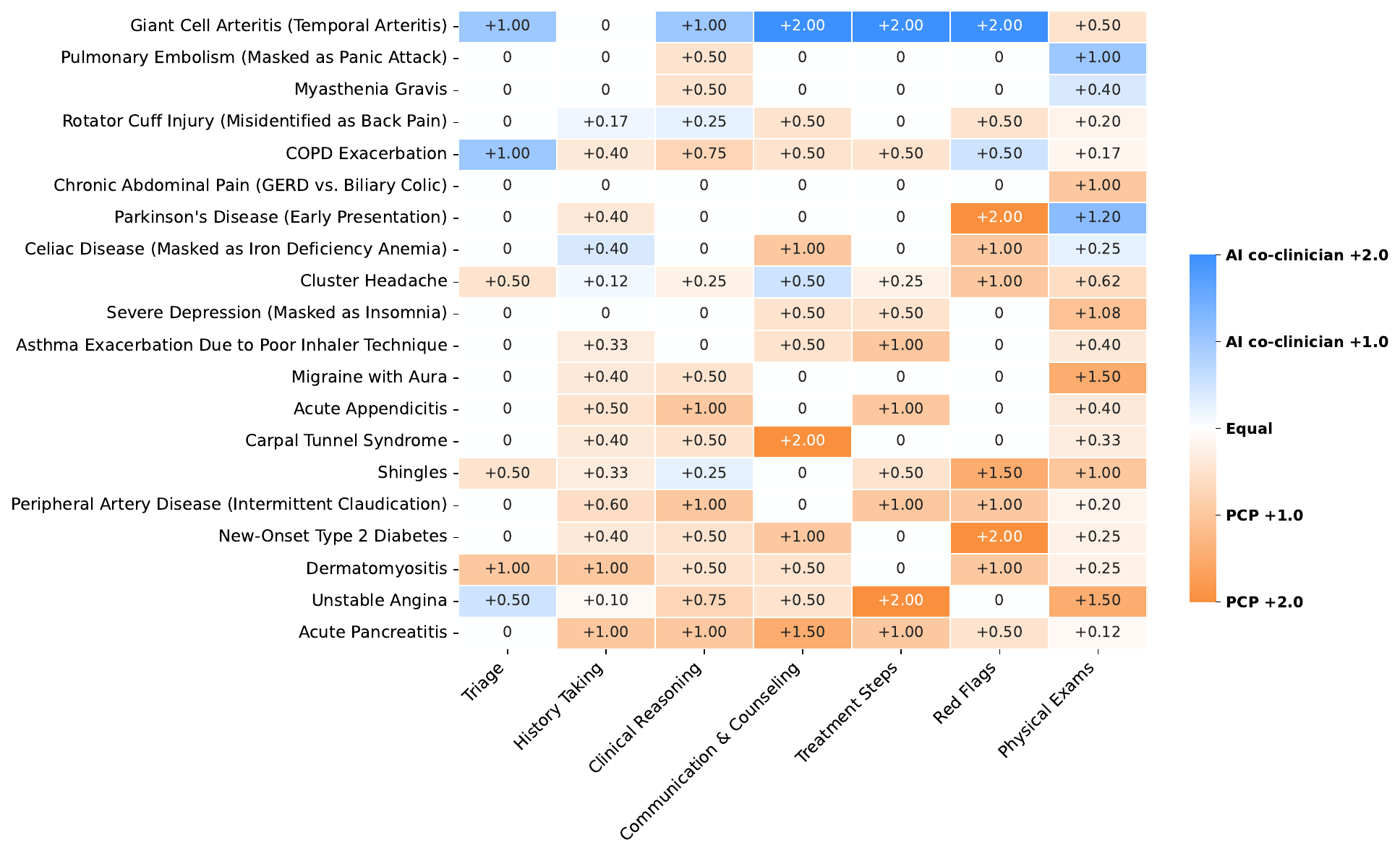}
    \caption{\footnotesize \textbf{Instance-level discrepancies: \MedAstraName{} vs. Primary Care Physicians.} Performance map showing the difference in average rubric scores on a 0-2 scale (0=poor, 2=excellent) for each scenario. \textbf{Blue} indicates the PCP scored higher, \textbf{Red} indicates \MedAstraName{} scored higher. While \MedAstraName{} achieves parity in History Taking for several respiratory and musculoskeletal cases, consistent deficits appear in the Physical Exam column (e.g., Unstable Angina, Shingles) and Red Flag detection for high-acuity scenarios.} 
    \label{fig:gap_map_pcp}
\end{figure}

Case-specific comparative analysis further clarified the performance discrepancies between study arms (Figure~\ref{fig:gap_map_pcp}). Differences in performance were not uniform, varying by both domain and condition. Across the 140 case-specific domain-level comparisons between \MedAstraName{} and primary care physicians, PCPs outperformed \MedAstraName{} in 72 instances, the two were equivalent in 49, and \MedAstraName{} outperformed PCPs in 19 instances. Triage was the clearest area of parity, whereas PCPs more consistently outperformed \MedAstraName{} in physical examination and treatment-related domains. 

Moreover, across the 140 case-specific domain-level comparisons between \MedAstraName{} and GPT-Realtime, \MedAstraName{} was superior in 100 instances, GPT-Realtime was superior in 8 and they were equivalent in 32 (Figure~\ref{fig:gap_map_gpt} in the Appendix). Areas where \MedAstraName{} demonstrates consistent advantages over GPT-Realtime were history-taking, clinical reasoning and physical examination.

\subsection{Effect of Clinical Planner Module}

\begin{figure}[htbp]
    \centering
    
    \begin{subfigure}[b]{\textwidth}
        \caption{Case-specific evaluation rubric} 
        \centering
        \includegraphics[width=\linewidth]{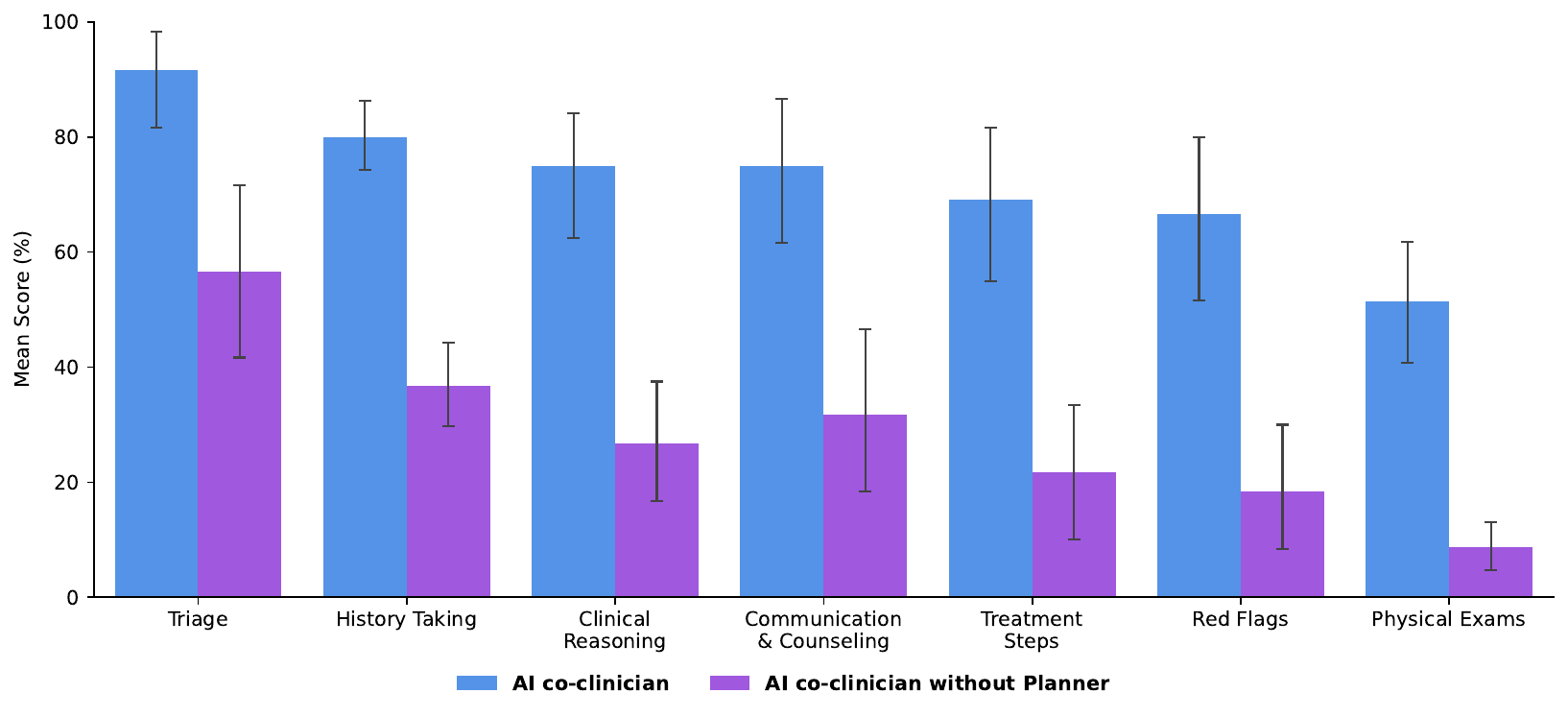}
    \end{subfigure}
    \begin{subfigure}[b]{\textwidth}
        \caption{Universal evaluation rubric}
        \centering
        \includegraphics[width=\linewidth]{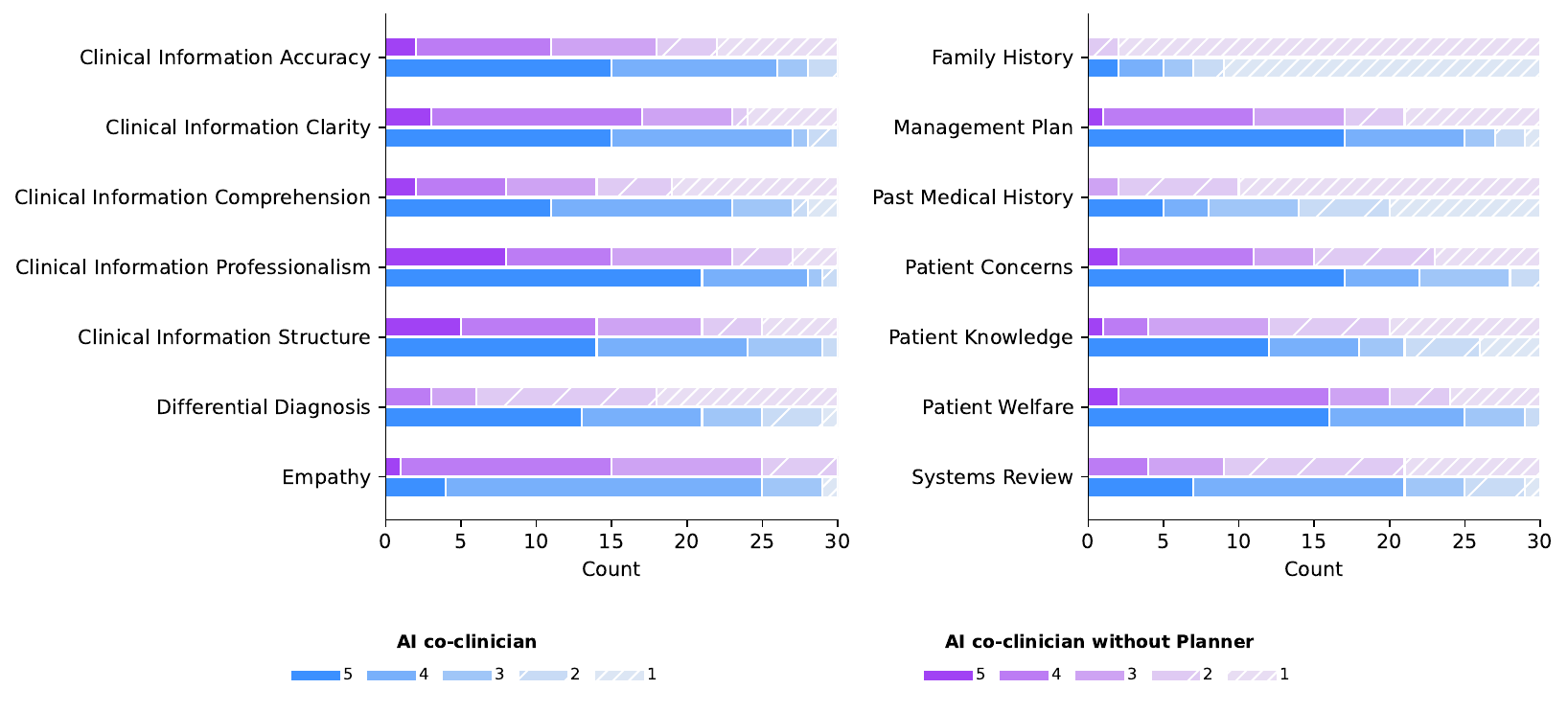}
    \end{subfigure}
    
    \caption{\footnotesize \textbf{Ablation of Clinical Planner in \MedAstraName{}.} 
    \textbf{(a) Case-specific evaluation rubric.} Scenario-tailored criteria assessing seven domains: history-taking completeness, audiovisual cue extraction, patient-assisted physical examination (guided maneuvers and interpretation), clinical reasoning accuracy, communication and counseling, treatment recommendations, triage disposition, and red-flag detection. Each item uses anchored 0--2 scoring to distinguish omissions, partial completion, and fully appropriate performance. Error bars correspond to 95\% confidence intervals. Statistical significance ($* = p < 0.05$) was evaluated using ordinary least squares regression with fixed effects for clinical scenario and patient actor.
    \textbf{(b) Universal evaluation rubric.} Domain-agnostic criteria applied across all encounters, assessing communication quality (clarity, empathy, rapport), clinical reasoning process (differential generation, information synthesis), safety behaviors (red-flag screening, appropriate escalation), and overall consultation structure independent of case-specific content. Ratings are based on a 1--5 Likert scale, where a score of 5 represents optimal performance (indicated by full color fill) and a score of 1 represents poor performance (indicated by hatched patterns).}
    \label{fig:universal_rubric_planner}
\end{figure}

The inclusion of the clinical planner module had significant impact on performance (see Fig. \ref{fig:universal_rubric_planner}). \MedAstraName{} outperformed the ablation arm in every universal-rubric and case-specific domain. The largest gains were observed in case-specific history taking, clinical reasoning, red-flag detection, and physical examination, supporting the importance of supervisory planning for medically coherent real-time interaction. Qualitative feedback from patient actors suggests that the lack of the higher-level planning resulted in relatively ``quicker encounters'', with a stronger tendency for ``deferring most of the workup and differential to in-person doctors'', which may explain the lower scores across the board for this simpler version of the system.

\subsection{Qualitative failure analysis}

All 20 \MedAstraName{} videos and transcripts were reviewed by study team members. Directed content analysis of physical exam domains showed clusters of failure modes for \MedAstraName{}. These included omitting feasible examination maneuvers, providing insufficiently specific instructions to guide the patient, and drawing conclusions without observable confirmation—reflecting a causal chain (maneuver selection $\rightarrow$ stepwise instruction $\rightarrow$ real-time adaptation $\rightarrow$ visual interpretation). We also saw nuanced discrepancies between TelePACES and case-specific rubrics, most notable in red flag symptoms, but also in treatment steps, where agent recommendations were more frequently incomplete or less specific than PCPs'. We also found dissociation between AI systems' conversational fluency and clinical performance. The system maintained a highly professional and articulate bedside manner while also being capable of omitting nuanced history-taking questions in specific conditions, a finding that was not seen among our clinicians, where conversational fluency fluctuated with condition-specific confidence. A novel finding from our analysis was ``contextual completion'', the capacity for the agent to infer examination findings or clinical conclusions from the expected trajectory of the case rather than from directly observable evidence. For example, the agent sometimes presented confident physical examination interpretations (e.g., reporting specific auscultatory findings) that were never actually elicited, observed, or confirmable in the interaction. To our knowledge, this represents a new safety risk for clinical AI systems, specific to telemedical interactions. 

\subsection{Inter-rater reliability}
Measures of inter-rater reliability are shown in Table~\ref{tab:irr}. The case-specific rubrics showed moderate-to-high rank preservation, with Kendall’s Tau-b values ranging from 0.545 to 0.915. The highest agreement was seen for total rubric score, history taking, red flags, and physical examination. Conversely the universal-rubric domains from TelePACES were more variable. While some axes showed moderate to high concordance such as family history, past medical history, differential diagnosis, management plans and clinical information accuracy, several others that could be considered more subjective (such as explaining information clearly and addressing patient concerns) had poor reliability. These findings suggest that the narrower and more behaviorally anchored case-specific rubric items were more reproducible across replicated encounters than several of the broader global ratings. 

\begin{table}[!htbp]
    \footnotesize
    \centering
    \small
    \caption{\textbf{Rank preservation per evaluation category.} For the 10 scenarios which were completed by two patient actors, Kendall's Tau-b is computed for each category to assess the inter-rater reliability for rankings between PCPs, \MedAstraName{}, and GPT-Realtime. For the rubric categories, we used the percent of points earned in each category, while for the TelePACES categories we used the numerical Likert ratings.}
    \label{tab:irr}
    \begin{tabularx}{\textwidth}{l X r}
        \toprule
        \textbf{Evaluation} & \textbf{Category} & \textbf{Kendall's Tau-b} \\
        \midrule
        
        \multirow{8}{*}{\textbf{Rubric}} 
        & Total Rubric Score & 0.915 \\
        & History Taking & 0.878 \\
        & Red Flags & 0.713 \\
        & Physical Exams & 0.712 \\
        & Communication \& Counseling & 0.609 \\
        & Treatment Steps & 0.590 \\
        & Clinical Reasoning & 0.552 \\
        & Triage & 0.545 \\
        
        \midrule 
        
        \multirow{14}{*}{\textbf{TelePACES}} 
        & To what extent did the doctor elicit the FAMILY HISTORY? & 0.862 \\
        & To what extent did the doctor elicit the PAST MEDICAL HISTORY? & 0.728 \\
        & To what extent did the doctor construct a sensible DIFFERENTIAL DIAGNOSIS? & 0.710 \\
        & To what extent did the doctor explain relevant clinical information ACCURATELY? & 0.657 \\
        & To what extent did the doctor select a comprehensive, sensible and appropriate MANAGEMENT PLAN? & 0.641 \\
        & To what extent did the doctor elicit the SYSTEMS REVIEW? & 0.573 \\
        & To what extent did the doctor confirm the patient's knowledge and understanding? & 0.548 \\
        & How empathic was the doctor? & 0.438 \\
        & To what extent did the doctor explain relevant clinical information PROFESSIONALLY? & 0.352 \\
        & To what extent did the doctor maintain the patient's welfare? & 0.352 \\
        & To what extent did the doctor explain relevant clinical information COMPREHENSIVELY? & 0.331 \\
        & To what extent did the doctor explain relevant clinical information WITH STRUCTURE? & 0.227 \\
        & To what extent did the doctor explain relevant clinical information CLEARLY? & 0.144 \\
        & To what extent did the doctor seek, detect, acknowledge and attempt to address the patient's concerns? & -0.063 \\
        
        \bottomrule
    \end{tabularx}
\end{table}

\section{Discussion}

In this work, we introduce \MedAstraName{}, a real-time audiovisual conversational medical agent that extends conversational AI capabilities to live video-based clinical encounters. We developed a new case-specific evaluation framework to enable assessment of audiovisual AI telehealth agents, specifically designed to identify nuanced audiovisual and safety features that traditional global ratings approaches might omit. We tested this new evaluation framework in a prospective telehealth OSCE study across two institutions with \MedAstraName{} and GPT-realtime, as well as an ablative arm (\MedAstraName{}-without-Planner), with human PCPs as a comparator.  We showed that \MedAstraName{} approached human performance in diagnostic accuracy, an observation consistent with text-based systems, showing parity with human clinicians in text-based experimental or clinical deployment environments \citep{Tu2025, zeltzer2025comparison}.  We found clinician-approximating performance in triage, an encouraging finding given that recent studies have suggested relatively poorer performance from text-based systems on triage tasks under certain constraints~\citep{Ramaswamy2026-kz}. 

A striking aspect of these results is that a real-time multimodal system was able to conduct any patient-assisted physical examinations at all. The system was able to propose examination steps, deliver instructions in natural language, adapt to the patient’s responses, and incorporate visual and auditory cues into its evolving assessment: a capability that, until recently, was largely outside the scope of text-only clinical AI. Our ablation study suggests that our dual-agent architecture is responsible for these performance gains. Even when incomplete, this represents a meaningful shift: the model is not merely generating an exam description or relying on patient textual input, but participating in an embodied, interactive measurement process over video. That this is now possible suggests that future multimodal assistants may ultimately help standardize elements of the virtual physical exam, improve the consistency of telehealth assessments, and extend basic examination capability to settings where clinician time or access is limited.  

Performance on both evoked and unevoked physical exam maneuvers illustrated the range of examination capabilities that a real-time multimodal system was able to achieve:

\begin{itemize}
    \item \textbf{Asthma (inhaler technique):} The model elicited a demonstration of inhaler use over video and identified incorrect technique—failure to shake the canister, holding the inhaler several inches from the mouth, and mistimed inhalation—then coached the patient through a corrected sequence that produced immediate, observable symptomatic relief.
    \item \textbf{Rotator cuff injury (guided provocative testing):} Moving beyond the patient's initial localization to ``upper back pain,'' the model operationalized a focused shoulder examination by guiding a supraspinatus ``empty-can'' maneuver, unmasking pain and weakness consistent with a rotator cuff injury.
    \item \textbf{Biliary pathology (remote bedside test):} The model approximated a classic bedside test remotely by instructing the patient to palpate the right upper quadrant during inspiration, reproducing inspiratory arrest consistent with a positive Murphy's sign.
    \item \textbf{Dermatomyositis (visual inspection and functional testing):} The model identified knuckle lesions consistent with Gottron's papules via direct visual inspection while having the patient demonstrate proximal weakness with overhead arm elevation and rising from a chair.
\end{itemize}

However, our evaluation also uncovered dissociation between case-specific and generic TelePACES rubrics. Despite strong apparent performance on differential diagnosis and management plan measures, \MedAstraName{} missed red flags and essential treatment steps more frequently than the equivalently-rated PCPs. Critical exam maneuvers were also more frequently omitted. Some of these examples include:

\begin{itemize}
    \item \textbf{Shingles (red flag):} While the model appropriately identified the dermatomal distribution of shingles, it did not ask questions to appropriately exclude infectious causes.
    \item \textbf{Myasthenia exam (examination):} The model appropriately attempted to guide the user through an extraocular movement exam, but instructed the patient actor to "follow my finger" (which the system does not have). 
    \item \textbf{Unstable angina (diagnostic next steps)}: The model routinely ended the conversation early and insisted upon the patient going to the emergency room. However, it would not give discrete recommendations for diagnostic next steps, for example EKG, serial troponins, or coronary angiogram.
    \item \textbf{Pancreatitis (physical exam)}: The patient actor gestured to their umbilical region, but the model interpreted the epigastric region, which affected its further questions. It also  inferred rebound tenderness on exam without further clarifying questions. 
    \item \textbf{Depression (red flag)}: The system did not appropriately screen for self harm in the depression case.
\end{itemize}

Our study has important implications for the rapidly-changing clinical landscape of patient-facing AI systems. Text-based approaches to augmenting the patient's clinical encounter have seen rapid exploration while evaluation of their performance based on global rubrics have suggested equivalence, or even superiority, to human clinicians in comparative studies where consultation is restricted to text-chat \citep{zeltzer2025comparison, Tu2025}. The assumption that performance on global rubrics is sufficient may have been well grounded as validation studies in medical education have suggested that these rubrics have better inter-case reliability, better construct validity, and better concurrent validity than case-specific rubrics~\citep{Regehr1998-ai}. However, the discordance seen in our study raises an important observation that AI systems are not humans, so an over-reliance on quality measures developed and validated for human medical students being observed by skilled and trained medical educators might not be suitable for exposing critical safety failures of AI systems. 

For example, a patient with a COPD exacerbation communicating with a text-based chatbot system may never volunteer that they are critically short of breath or using accessory muscles to breathe. A global evaluation of that text-chat might suggest strong performance, especially if the system is respectful and elicits appropriate historical findings. However, if the unevoked contextual case-specific factors are not recognized, which would easily be perceived in a human telemedical encounter, such a patient diverted to AI text-chat alone could suffer negative health consequences. Fortunately, most clinical deployments operate in a strict human-in-the-loop system, where text-based chatbots serve as initial pre-visit assistive tools prior to telehealth clinician encounters. But these findings raise important cautions about human-on-the-loop approaches or the path to fully autonomous "AI clinician" systems and highlight that further deployments need to be carefully evaluated for the contextual factors that are missed by global rubrics.

While \MedAstraName{} represents a clear step forward in patient-facing AI agents, there remains an uneven or "jagged" technological frontier for system performance (\cite{DellAcquaEtAl2023JaggedFrontier}). Within our current evaluation framework, examination, especially of unevoked signs, lagged considerably behind other abilities. Furthermore, communication and counseling scores may have been artificially inflated given the simulated nature of this study. Data from our recent prospective clinical trial of the articulate medical intelligence explorer (AMIE) system suggested that patients in clinical care rated chatbot empathy considerably lower than patient actors, who are not discussing their own health \citep{brodeur2026prospectiveclinicalfeasibilitystudy}. Furthermore, our case-specific rubrics do not wholly define the frontier of performance in telemedicine. Prior evaluation work on management reasoning shows that defining appropriate management decisions is considerably more difficult in the real world than in constructed cases \citep{wu2025firstnoharmclinicallysafe, Goh2025-kt}. Management reasoning is highly context specific and relies on patient values, health system constraints, and local practice, among many other factors \citep{Cook2023-zq}. These include "intangibles" such as appropriate humility under uncertainty, consistency of priorities, recognition of limits, and the ability to signal when escalation is necessary. \citep{goddard2012automation}. Even the act of diagnosis, which theoretically has a "correct" answer, is subject to the same factors in the experimental literature \citep{McBee2017-vp} and may be considerably harder to evaluate in real-world settings where a final diagnosis is not always objectively apparent. More nuanced study is necessary to better define the performance of agents on this uneven technical frontier before such agents can be used without safety measures such as a strict human-in-the-loop system.

For example, a novel failure mode not previously described in the literature and specific to the task of real-time audio-video clinical interactions was “contextual completion”, in which the model inferred exam findings from the scenario trajectory or diagnostic priors rather than from observable evidence. This novel observation is particularly notable because it could propagate inaccurate conclusions about the physical examination: models might present confident exam interpretations that were not actually elicited, observed, or confirmable in the interaction. In practice, this would risk converting plausibility into overconfidence \citep{pal2023med, asadi2026mirage} especially in cases where safety hinged on subtle findings. It could also more directly be a cause for misdiagnosis or errors of investigation and treatment. The promise of agents for healthcare lies partly in the ability for scalable use under physician supervision. The detection and correction of erroneous "contextual completions" represents an important area for further work and rigorous safety evaluations.

\subsection{Study Limitations}
Our study has several important limitations regarding scenario design, participant demographics, our user interface, evaluation instruments and scope of technical comparisons. The clinical cases in this study were intentionally selected to feature presentations where specific audiovisual cues are critical to accurate diagnosis. Because of this targeted selection, the distribution of scenarios may have disproportionately emphasized settings where a PCP or multimodal model would naturally outperform a text-only AI system, and we did not compare performance in this study to text-chat AI systems. 

The encounters were simulated and standardized patients, even when well trained, cannot fully represent the variability of real-world telehealth including differences in digital literacy, home environments, camera quality, and patient behavior. Furthermore, our study design did not encompass the widespread real-world variation in disease severity or the diverse clinical presentations seen across early versus late stages of pathology. Because the patient actors did not possess true underlying pathology, representations of physical pain, distress, or signs (e.g. dermatological lesions) may not have replicated the full realism of a true clinical encounter. Patients were role-played by medically-trained residents who would respond to guided physical exams with great precision and likely appropriate clinical vocabulary. The study also did not examine patients in whom the presenting complaint was a self-resolving transitory symptom with no underlying "ground-truth" disease process, which represent an important control group to observe AI systems' calibration.

A limited demographic range of participating patient actors prevented meaningful analysis of accessibility or the impact of demographic factors such as gender, age, and sociocultural background. These are known to have impact in care delivery influencing patient trust and the effective utilization of telemedicine. Consultations were only performed in English and we did not specifically examine challenges of equity, bias, language, socio-economic context or cultural expectations of the healthcare encounter which remain essential for real-world assessment.

The assessment utilized a strictly interface-blinded design. In all study arms, users heard the spoken response and saw a text transcript generated from the video stream, but they did not see a visual representation of the AI agent (for example rendered as an avatar) or the physician's face. However, measuring the impact of such visual perception adds variables that require rigorous further research which we did not explore. Within the interface-blinded design we did not sensitivity-test for differences in the tone, expressivity, or specific persona of the AI models, which may also significantly influence perceived empathy and clinical rapport, nor examine longitudinal usage and adaptation of models or patients to rapport over time.

PCP performance in this study may overestimate routine clinical performance due to the \textit{Hawthorne effect} \citep{mccarney2007hawthorne}. Physicians who know they are being observed and assessed tend to perform significantly closer to their best possible practice than they would in unobserved routine consultations \citep{Goodwin2017-at}. Furthermore, research using incognito or unannounced standardized patients has documented substantive gaps in physician performance in real-world settings that are not apparent when clinicians know they are being assessed \citep{Schwartz2021-kp, Rethans2007-xu}. The PCP arm represents a narrow sample of board-certified academic physicians at two tertiary centres. In reality, there is substantial variability in PCP skills, experience, and practice patterns among physicians in broader practice settings \citep{Sirovich2008-wg, Lipner2015-oe}.

\subsection{Recommendations for Future Evaluations}

Grading multimodal encounters is intrinsically harder than grading text: evaluators must track what was asked, what was done on video, what was actually observable, and what the system claimed to find. In this study, we relied on our patient actors, internal medicine residents, to grade the multimodal agents. Future evaluations should use multiple reviewers to grade assessments in order to establish appropriate reliability of rubrics. Human scoring would ideally be paired with structured, machine-readable trace data (time-stamped transcripts, action markers for exam maneuvers, and “evidence tags” linking conclusions to specific observed inputs), enabling adjudication of whether an exam finding was elicited and supported. This approach could also be used to develop autograder tools to extend evaluation. 

Our study protocol did not anticipate some experimental technology failure modes, such as agent timeouts or terminations, requiring us to repeat 5 of the 80 AI encounters. Standardized patient performance should be assured through professional patient-actors and auditing (including inter-actor consistency checks) to reduce variability in how prompts, affect, and responses are delivered across arms. Similarly, model and agentic harness behavior could be subjected to standard study participation instructions while minimizing cross-arm differences unrelated to modality. Importantly,future evaluation protocols should explicitly screen for unsupported physical-exam assertions and require agents to separate “observed,” “patient-reported,” and “inferred” findings; reducing hallucinated exam conclusions that arise from clinical-context inference rather than from verifiable evidence.

\subsection{Technical Challenges for Real-time Audio-Video Interactions} Because this study represented the first demonstration of a real-time AV medical agent, providing expanded details on its failure modes may be helpful for informing future work and helping to shape the direction of future research in this area:
\begin{itemize}
    \item \textbf{Contextual Completion:} Driven by underlying next-token prediction objectives, the model occasionally hallucinated physical examination findings based on clinical expectations rather than observable visual evidence. In live settings, converting statistical plausibility into visual overconfidence represents a novel safety risk that should be addressed.
    \item \textbf{Omission of `Unevoked' Red Flags}: The agent's lower performance in detecting "red flag" symptoms not explicitly mentioned by the patient represents a challenge in proactive clinical reasoning.
    \item \textbf{Environmental Variability:} Telehealth consultations occur in "uncalibrated" environments with high variability in camera quality, lighting, connectivity quality, and patient behavior. Handling environmental noise (like low light or mirrored camera views) reliably remains a foundational hurdle for safe deployment
\end{itemize}%

\subsection{Broader Implications}

These results mark a considerable shift in what is technically possible through the use of conversational diagnostic AI agents in telemedicine, with substantial progress towards accurate triage, diagnosis and treatment planning and history-taking in video interactions. The performance shortfalls between AI systems and physicians are less focused on conversational fluency, but rather occurred in clinically important domains such as reliability, grounding, clinically-accurate elicitation of symptoms and signs, and safety—particularly when clinical decisions depend on what is actually seen or heard rather than what is plausible from context. Addressing this shortfall will require evaluations that examine verifiable instances of carefully-recorded replicable physical examination, enforcing evidence-based examination routines, prevent unsupported inferences, and design safety behaviors that are consistent and testable.

\noindent\textbf{Extending Clinical Reach.} The ability to guide patient-assisted examination maneuvers, interpret audiovisual cues, and deliver structured clinical assessments in real time suggests possibly helpful future roles for these systems in assisting care or improving access. Even in well-resourced systems, standardizing portions of the virtual physical examination and improving the consistency of triage and escalation decisions could help reduce variability in telehealth quality. Significant progress is required before such systems could bridge toward autonomous use in diagnosing or treating real patients, and safety oversight is paramount. This research system and evaluation was designed to explore the technical feasibility of real-time multimodal clinical AI, operating under the strict assumption that future near-term utility might lie in clinician-supervised augmentation of the doctor-patient relationship, where diagnostic and therapeutic decisions require care provision by a licensed healthcare professional. An underappreciated advantage of AI-based systems is longitudinal memory. Large systematic reviews link greater personal continuity with reduced mortality and improved healthcare outcomes in primary care \citep{pereira2018continuity, engstrom2025personal}. Yet relational continuity has been steadily declining in many primary care systems, for example, continuity with a usual GP has fallen in English general practices over recent years \citep{levene2024ongoing, tammes2020continuity}, and broader literature has highlighted similar challenges in continuity and access in the United States. An AI system with access to the full history of prior interactions could maintain near-perfect informational continuity, a capability that may partially offset the trust gap observed in this study and that has no human equivalent at scale.

\noindent\textbf{Integration into Clinical Workflows.} Indeed, rather than functioning as autonomous replacements for clinicians, the current promising capabilities of these systems are more naturally positioned as structured collaborators: performing repeatable measurement tasks, communicating clearly with patients, generating summaries that might reduce cognitive load, and triggering timely handoffs when uncertainty or risk is high. The value of this integration will depend on whether systems can reliably separate observed findings from inferred conclusions, calibrate uncertainty, and maintain safety behaviors across diverse clinical contexts.

\section*{Acknowledgments}
This project was an extensive collaboration between many teams at Google DeepMind and Google Research. We thank Sumanth Dathathri for their comprehensive review and detailed feedback on the manuscript. We would also like to thank Andy Song, Reed Roberts, Mike Schaekermann and Tao Tu for their insightful discussions. We also thank Leen Verburgh, Rob Ashley, Armin Senoner and Adriana Fernandez Lara at Google DeepMind and Emma Moxhay, Simon Waldron and Matt Mager at Across the Pond for their contributions to the animations and visuals. We would like to thank SiWai Man and Gordon Turner for their operational support that enabled running this study smoothly. Lastly, we would like to thank Rishad Patel, Oriol Vinyals, Ed Chi, James Manyika and Rushika Fernandopulle for their support of this work.

\section*{Competing Interests}
This study was funded by Alphabet Inc and/or a subsidiary thereof (`G'). All authors are (or were) employees of Alphabet and may own stock as part of the standard compensation package.

\section*{Data Availability}
Further inquiries about our benchmarking procedures and data analysis may be addressed to the corresponding authors with a maximum response time of two weeks.

\section*{Code Availability}
Our system uses a Gemini family model as its base foundation model. Base Gemini models are generally available via Google Cloud APIs. However, the specific implementation relies on internal Google infrastructure and tooling, including Astra for agent orchestration. Due to this, and more importantly, the safety implications associated with the unmonitored deployment of AI systems in medical contexts, we are not open-sourcing the codebase or the specific prompts employed in our work at this time. In the interest of responsible innovation, we will be working with research partners, regulators, and healthcare providers to further validate and explore safe onward uses of our medical agents.

\clearpage
\bibliography{main}

\clearpage
\appendix
\input{appendix}

\end{document}

%% file: appendix.tex
\noindent \textbf{\LARGE{Appendix}}\\
\normalfont

\section{Rubric Details}

\begin{table}[!htbp]
    \centering
    \includegraphics[width=0.9\textwidth,keepaspectratio]{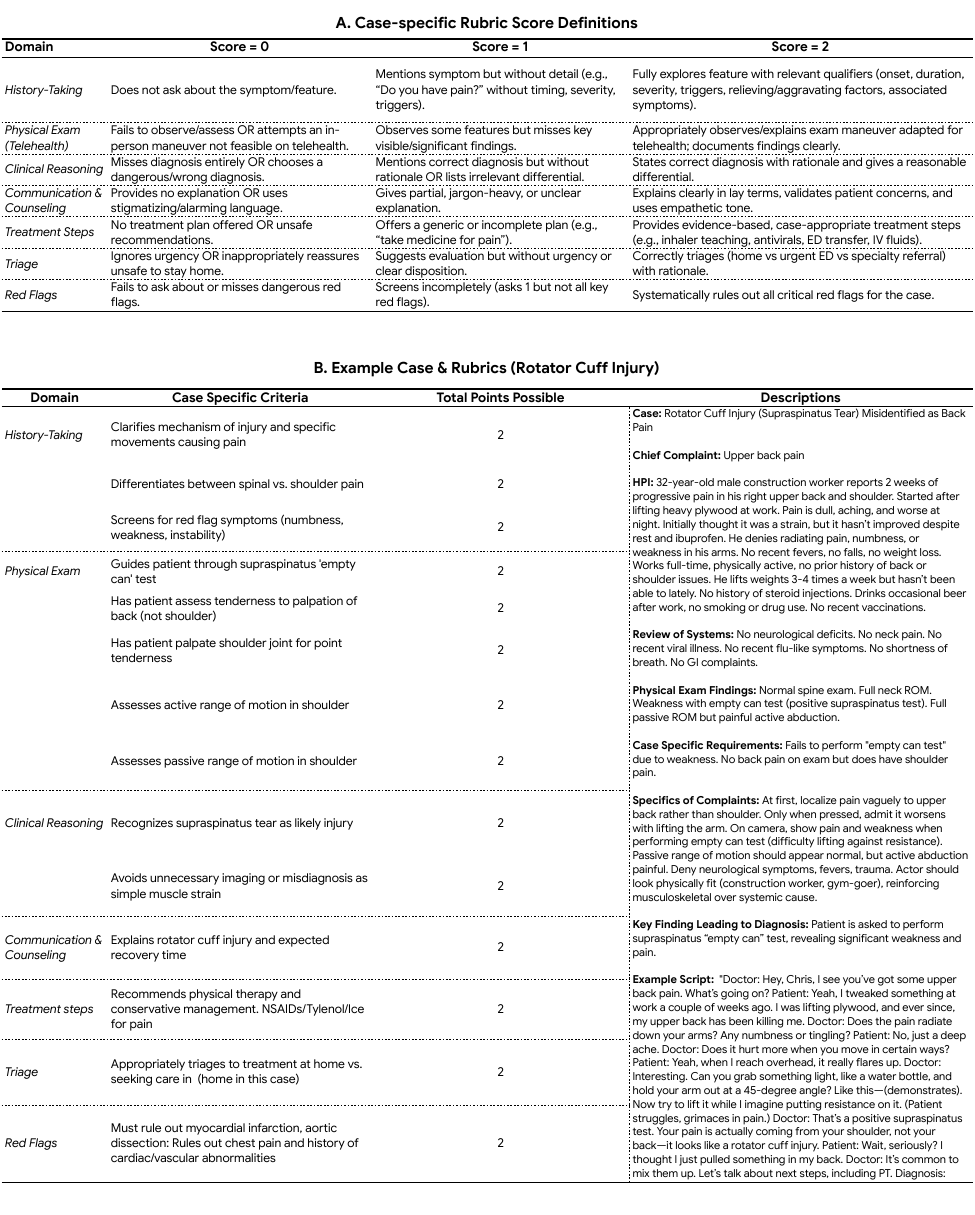}
    \caption{\footnotesize \textbf{Case-specific rubric domains and example items.} For each vignette, the case-specific rubric spans seven domains (History-Taking, Physical Exam [Telehealth], Clinical Reasoning, Communication and Counseling, Treatment Steps, Triage, and Red Flags), with items scored on a 0--2 scale.}
    \label{tab:case_specific_rubrics}
\end{table}

\begin{table}[!htbp]
    \centering
    \includegraphics[width=0.95\textwidth,height=0.84\textheight,keepaspectratio]{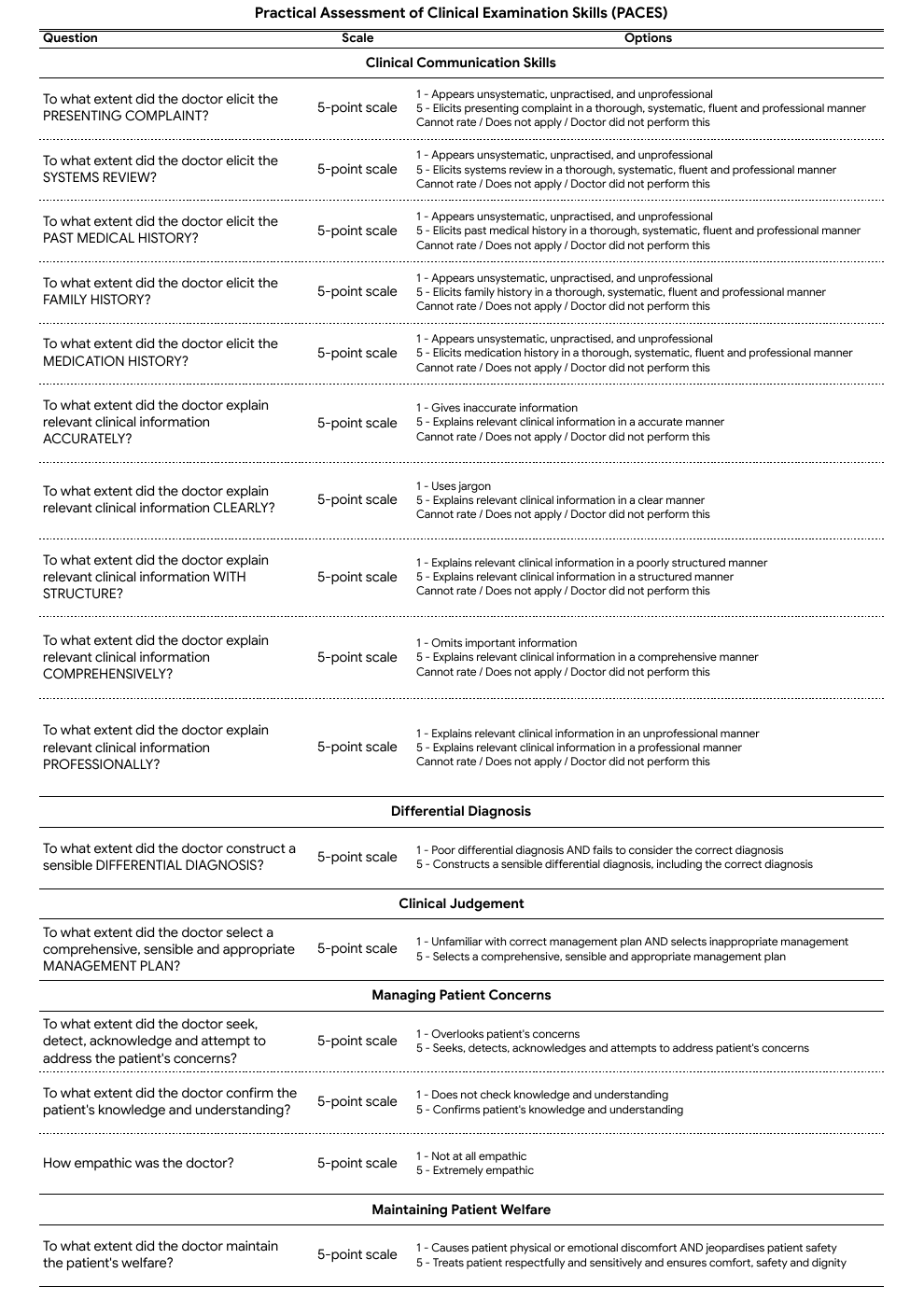}
    \caption{\footnotesize \textbf{Universal rubric domains and scoring.} The universal rubric, adapted from established assessment frameworks, Practical Assessment of Clinical Examination Skills (PACES), was applied across all vignettes to evaluate encounter-level performance consistently.}
    \label{tab:universal_rubrics}
\end{table}

\section{Detailed Comparison of \MedAstraName{} vs GPT-Realtime}

\begin{figure}[ht!]
    \centering
    \begin{subfigure}[a]{\textwidth}
        \centering
        \includegraphics[width=\linewidth]{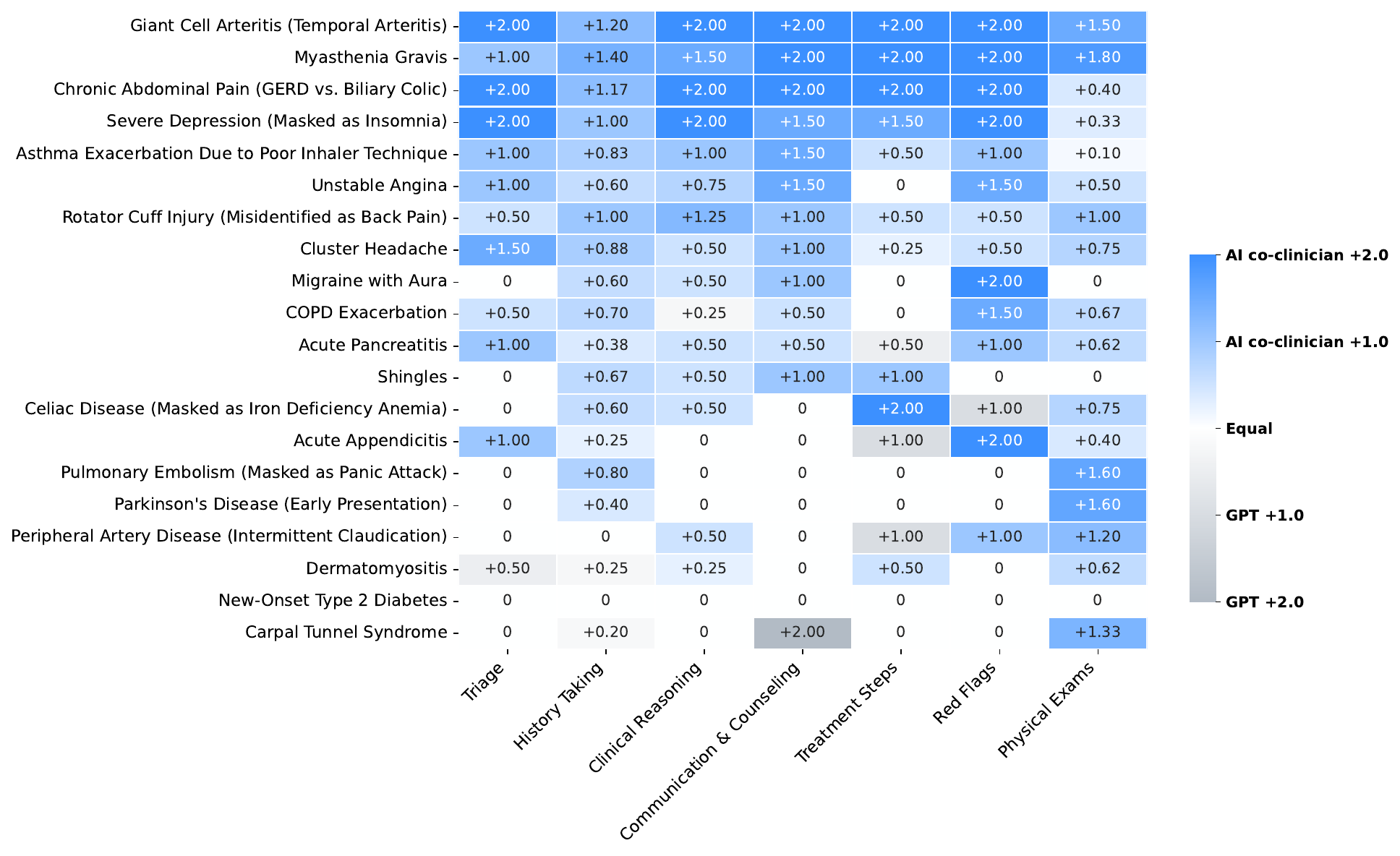}
    \end{subfigure}
    \begin{subfigure}[b]{\textwidth}
        \centering
        \includegraphics[width=0.935\linewidth]{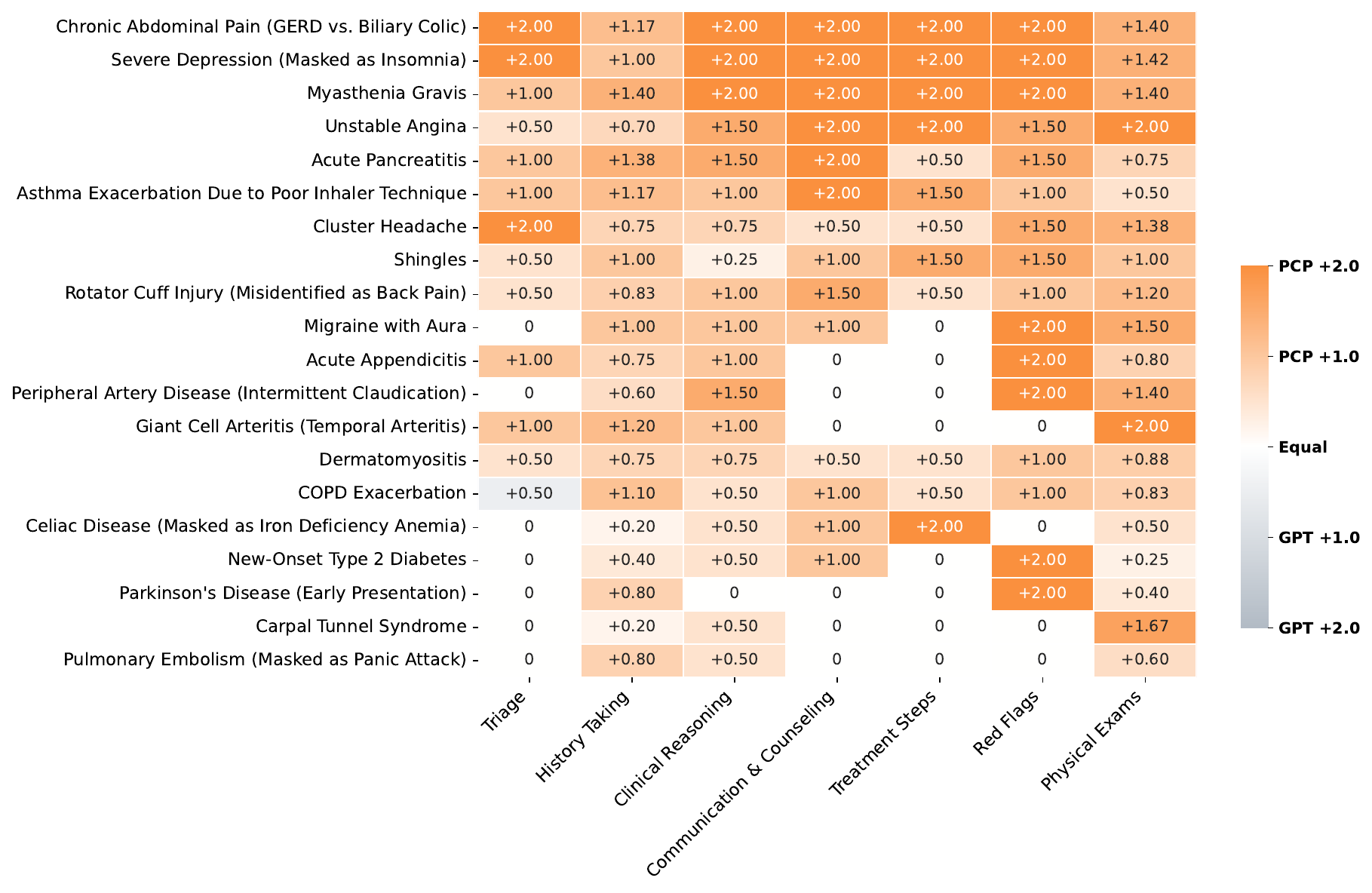}
    
    \end{subfigure}
    \vspace{-8mm}
    \caption{\footnotesize \textbf{Instance-level discrepancies with respect to GPT-Realtime} Performance map showing the difference in average rubric scores on a 0-2 scale (0=poor, 2=excellent) for each pair of scenario and domain (e.g., Triage). \textbf{Gray} indicates GPT-Realtime scored higher. \textbf{Blue} indicates the PCP scored higher, \textbf{Red} indicates \MedAstraName{} scored higher.}
    \label{fig:gap_map_gpt}
\end{figure}

\clearpage
\section{Additional Results}

\begin{table}[!hbtp]
    \centering
    \includegraphics[width=0.9\textwidth,keepaspectratio]{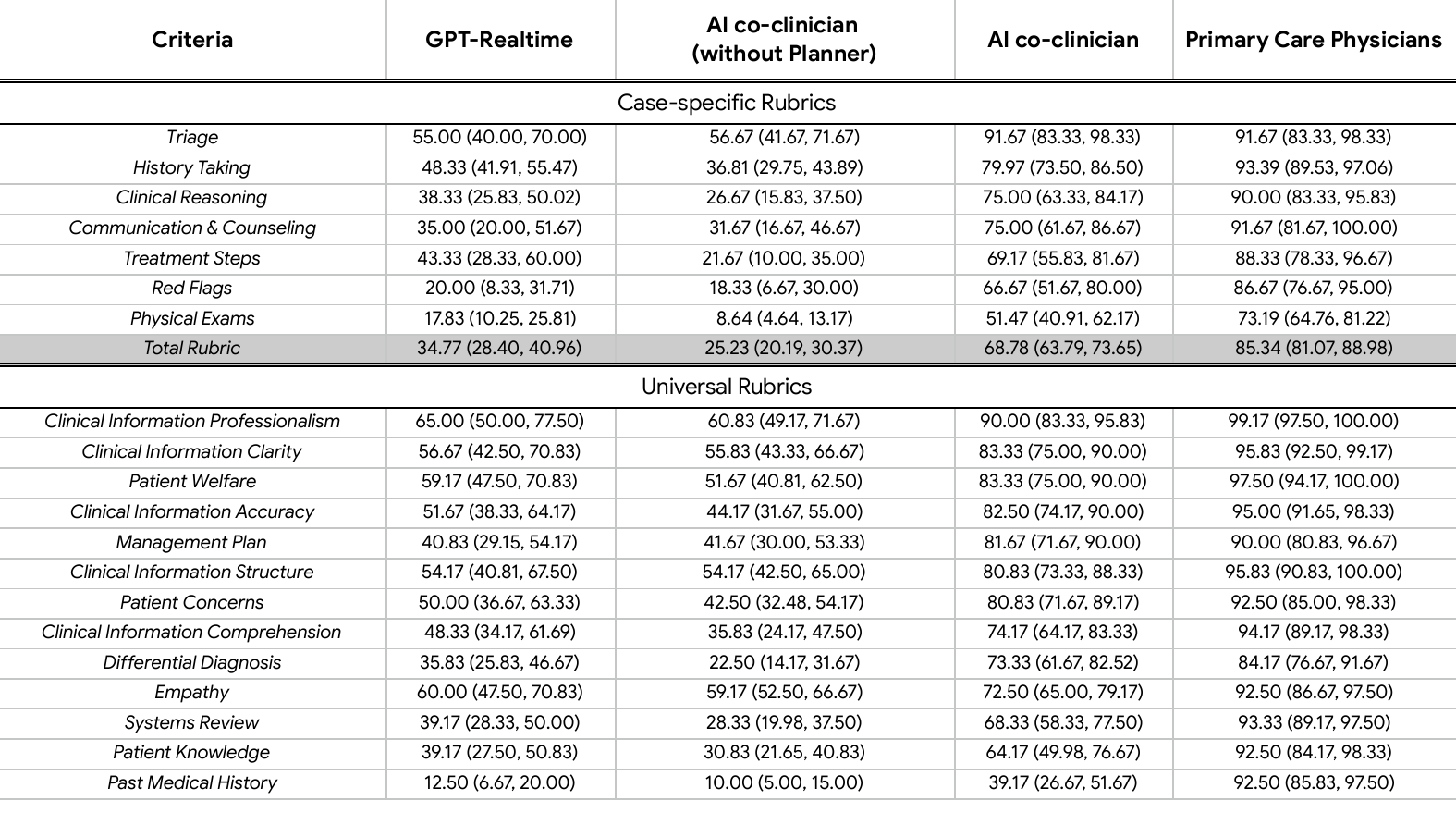}
    \caption{\footnotesize \textbf{Aggregated Scores of Universal and Case-Specific Rubrics.} This table provides the raw mean values and their corresponding confidence intervals (95\%) presented in Figure~\ref{fig:universal_rubric} and Figure~\ref{fig:domain_results}. Confidence intervals are estimated based on $10^4$ bootstrap samples.}
    \label{tab:percentage_scores_mean_and_ci}
\end{table}

\begin{figure}[!htbp]
    \centering
    \includegraphics[width=\textwidth]{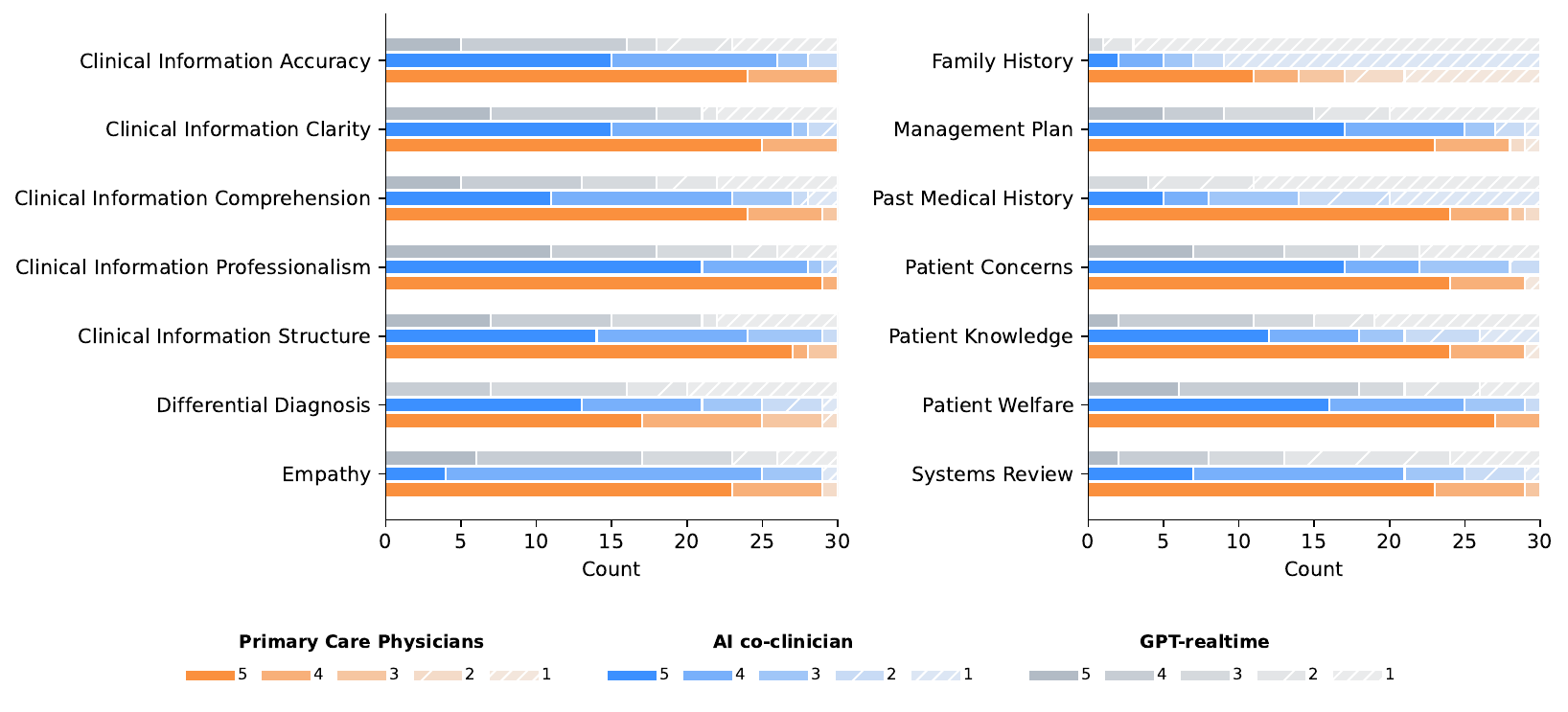}
    \caption{\footnotesize \textbf{Distribution of ratings on Universal Rubrics.} Domain-agnostic criteria applied across all encounters, assessing communication quality (clarity, empathy, rapport), clinical reasoning process (differential generation, information synthesis), safety behaviors (red-flag screening, appropriate escalation), and overall consultation structure independent of case-specific content. Ratings are based on a 1--5 Likert scale, where a score of 5 represents optimal performance (indicated by full color fill) and a score of 1 represents poor performance (indicated by hatched patterns).}
    \label{fig:universal_rubric_distribution}
\end{figure}

\clearpage
\section{Qualitative Analysis: Planner in Action - Myasthenia Gravis Case}
\label{section:planner_examples}
\begin{figure}[!htbp]
    \centering
    \includegraphics[width=\textwidth]{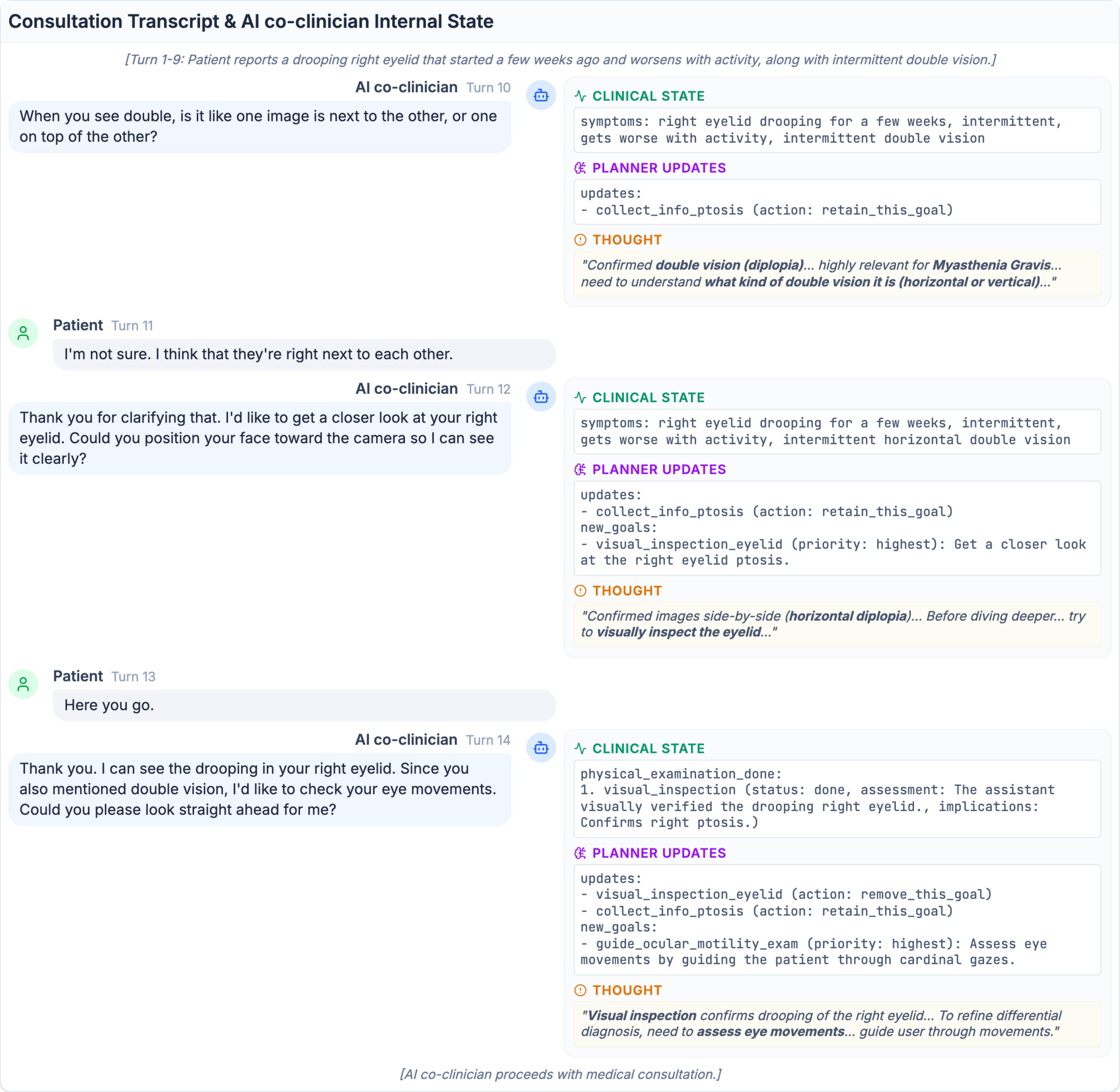}
    \caption{\footnotesize \textbf{Symptom-Directed Visual Assessment.} Stateless LLMs often become trapped in prolonged history-taking loops. In this example, once the patient confirms a critical subjective symptom (horizontal diplopia in Turn 11), the planner recognizes the optimal moment to seek visual evidence and dynamically injects a \texttt{visual\_inspection} goal. This state update prompts the agent to pause further questioning and immediately initiate a symptom-directed visual assessment, successfully corroborating the subjective history with objective clinical signs.}
    \label{fig:ai_co_clinician_visual_inspection}
\end{figure}

\begin{figure}[!htbp]
    \centering
    \includegraphics[width=\textwidth]{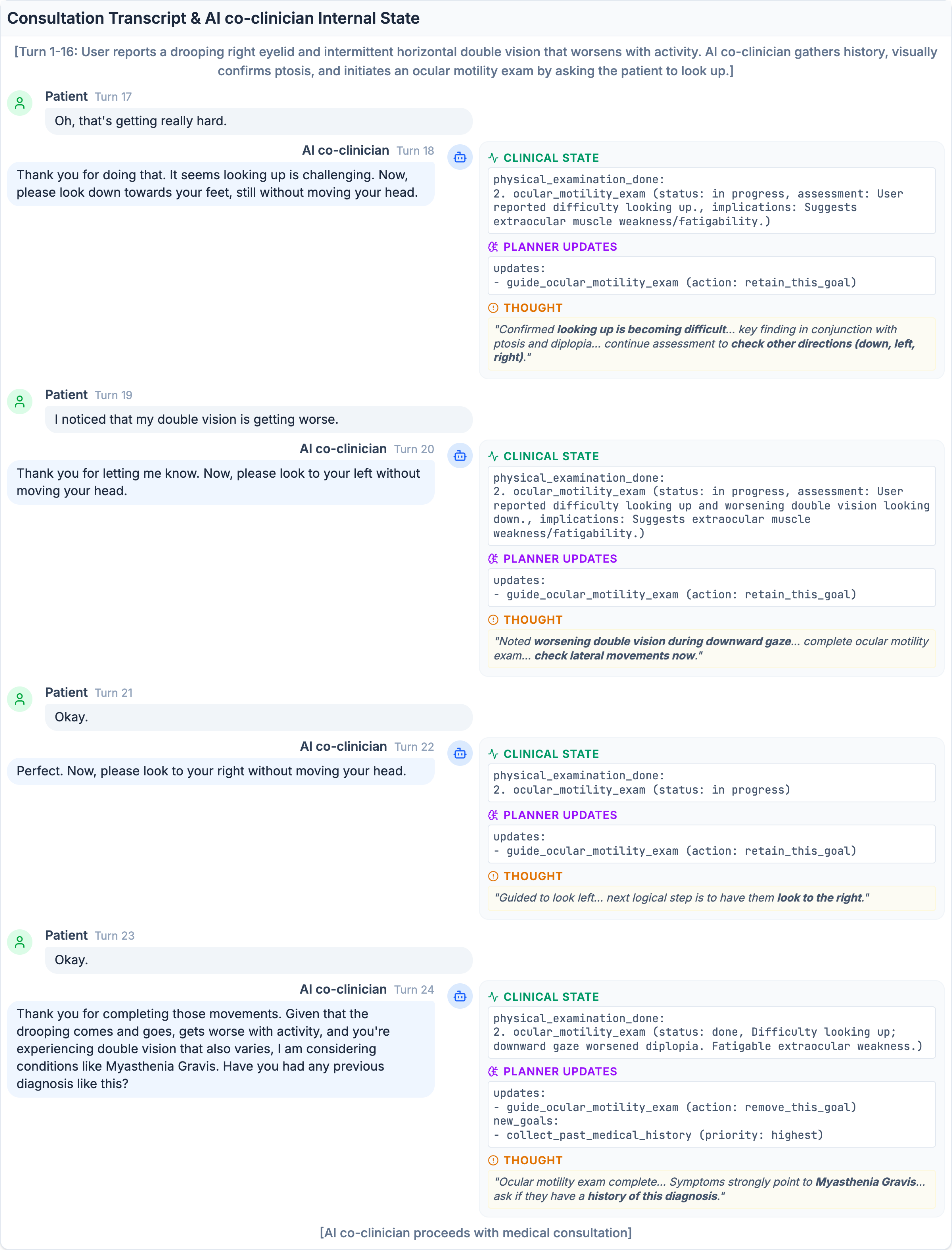}
    \caption{\footnotesize \textbf{Multi-Step Procedure Orchestration.} To conduct an accurate extraocular movement exam, instructions must be given sequentially rather than simultaneously. The planner retains the \texttt{guide\_ocular\_motility\_exam} goal across multiple turns (Turns 16-22). This allows the agent to methodically step through each cardinal gaze, logging distinct positive findings (fatigability in Turn 17, worsening diplopia in Turn 19) without losing track of the overall procedure.}
    \label{fig:ai_co_clinician_ocular_mobility}
\end{figure}

\begin{figure}[!htbp]
    \centering
    \includegraphics[width=\textwidth]{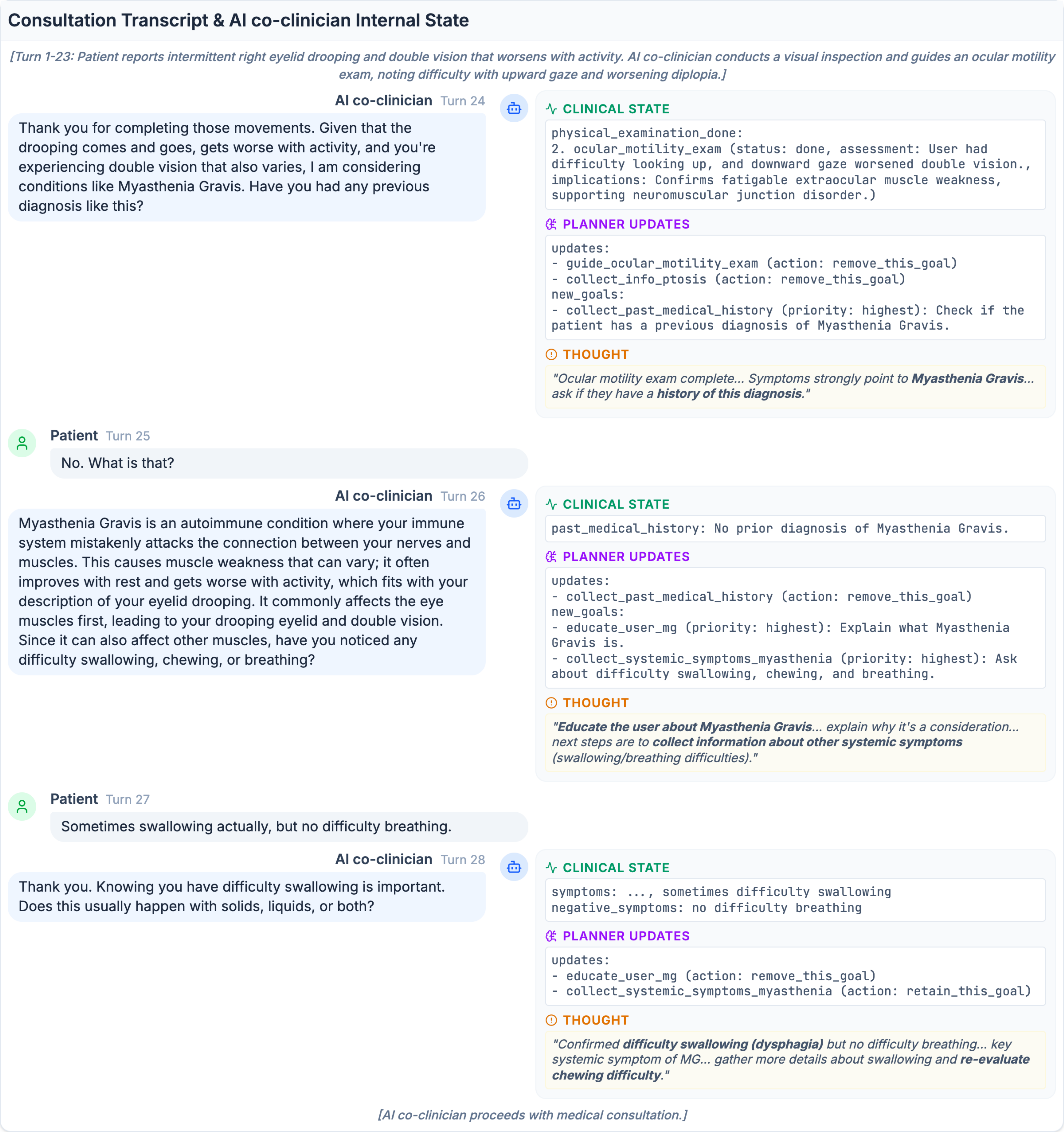}
    \caption{\footnotesize \textbf{Structured Context Switching.} When the patient interrupts the diagnostic workflow to ask about a disease (Turn 25), the planner dynamically injects a \texttt{educate\_user} goal to provide information to the patient while queuing the next clinical assessment. Once the patient's question is answered, the agent seamlessly returns to the diagnostic checklist, ensuring a thorough and patient-centric consultation.}
    \label{fig:ai_co_clinician_context_switch}
\end{figure}

\begin{figure}[hbt!]
    \centering
    \includegraphics[width=\textwidth]{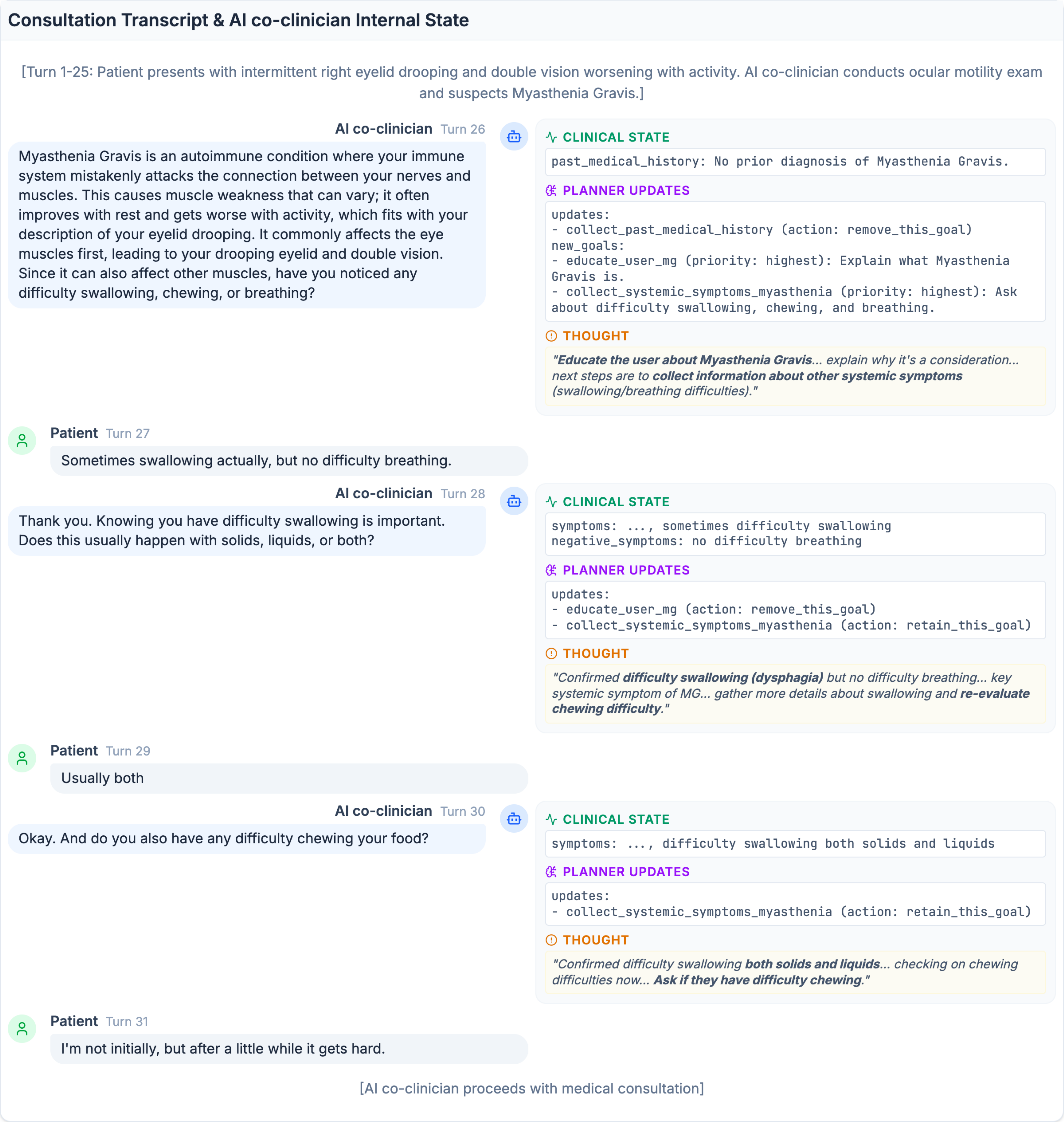}
    \caption{\footnotesize \textbf{Goal Retention for Incomplete Symptom Capture.} In Turn 26, the agent asks a compound question. The patient omits information regarding chewing difficulties (Turn 27). The planner retains the active symptom-collection goal across multiple conversational turns, enabling the agent to characterize the reported symptoms and successfully loop back to capture the omitted symptom (Turn 30).}
    \label{fig:ai_co_clinician_symptoms}
\end{figure}

\begin{figure}[hbt!]
    \centering
    \includegraphics[width=\textwidth]{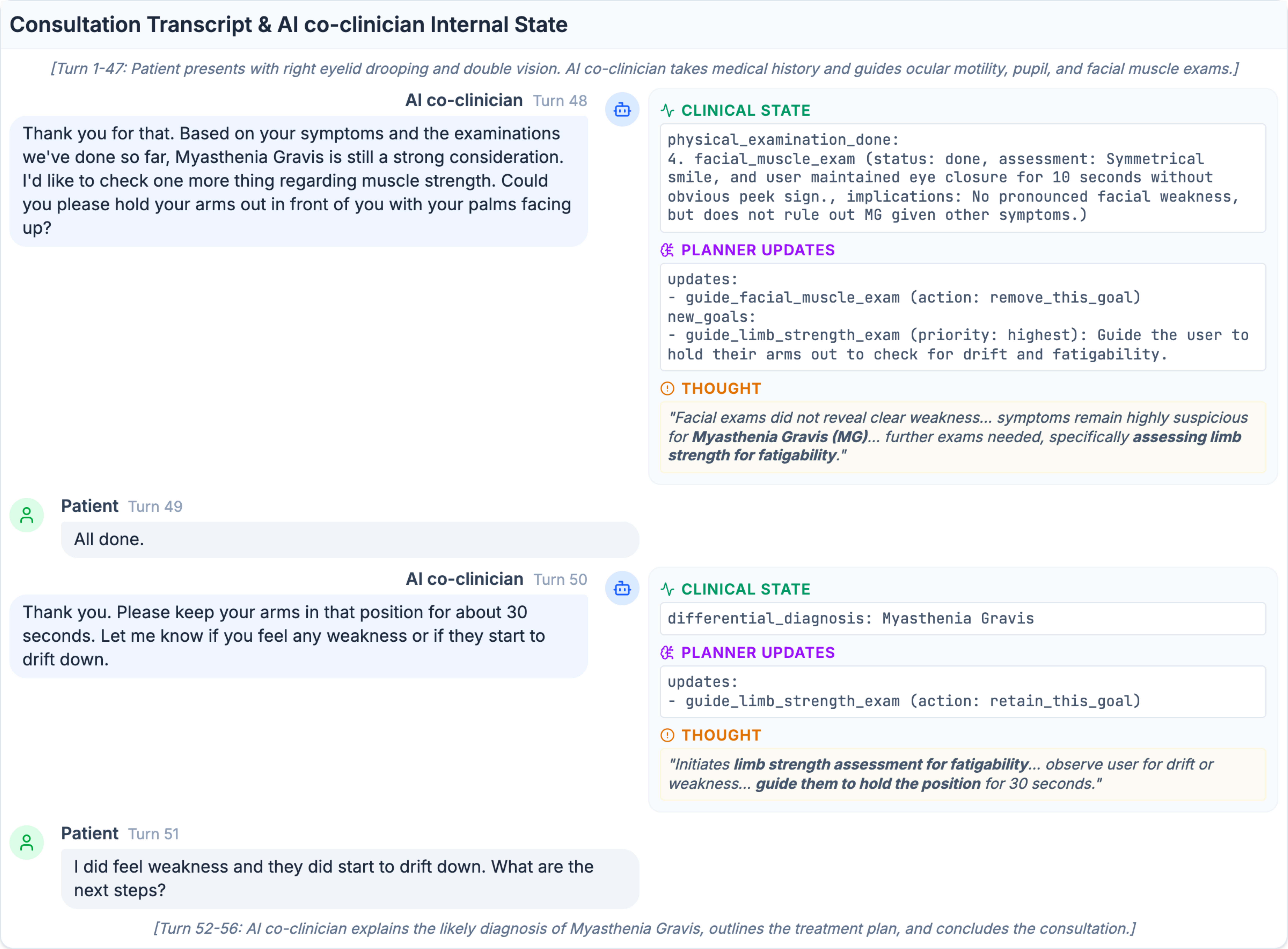}
    \caption{\footnotesize \textbf{Course-Correcting Guided Examination.} When testing for a condition characterized by fatigability (Myasthenia Gravis), a planner understands that brief maneuver is insufficient. When the patient briefly extends their arms (Turn 49), the planner retains the goal, forcing the agent to introduce a 30-second time constraint (Turn 50). This dynamic goal retention captures a critical positive clinical finding (drift) that a stateless model would have missed.}
    \label{fig:ai_co_clinician_limb_strength}
\end{figure}

To demonstrate the impact of planner's dynamic goal management on consultation quality, we analyze one of the study sessions involving a patient presenting with Myasthenia Gravis (MG). The following excerpts highlight how the planner guides the agent through complex clinical scenarios that stateless LLMs typically fail to navigate, such as confirming subjective symptoms via objective signs (Figure~\ref{fig:ai_co_clinician_visual_inspection}), orchestrating multi-step exams (Figure~\ref{fig:ai_co_clinician_ocular_mobility}), switching context dynamically (Figure~\ref{fig:ai_co_clinician_context_switch}), maintaining goal persistence (Figure~\ref{fig:ai_co_clinician_symptoms}), and course-correcting patient behaviors (Figure~\ref{fig:ai_co_clinician_limb_strength}).

\clearpage
\section{Demographics of Study Participants}

\begin{table}[!htbp]
    \centering
    \begin{tabular}{lc}
    \toprule
    Characteristic & Overall (N=10) \\
    \midrule
    Age, yr & 30.9 $\pm$ 1.7 \\
    Sex, no. (\%) &  \\
    \quad Male & 8 (80) \\
    \quad Female & 2 (20) \\
    Self-reported ethnicity, no. (\%) &  \\
    \quad White & 7 (70) \\
    \quad Asian & 3 (30) \\
    \bottomrule
    \end{tabular}
    \caption{\footnotesize \textbf{Demographic characteristics of study participants}}
    \label{tab:resident_demographics}
\end{table}